\documentclass[11pt]{article}

\PassOptionsToPackage{dvipsnames,table}{xcolor}

\usepackage[final]{acl}

\usepackage{times}
\usepackage{latexsym}
\usepackage[T1]{fontenc}
\usepackage[utf8]{inputenc}
\usepackage{microtype}
\usepackage{inconsolata}

\usepackage{graphicx}
\usepackage{subcaption}
\usepackage{booktabs}
\usepackage{array}
\usepackage{wrapfig}
\usepackage{float}
\usepackage{placeins}
\usepackage{afterpage}
\usepackage{multirow}
\usepackage{enumitem}
\usepackage{xcolor}
\usepackage{colortbl}
\usepackage[normalem]{ulem}
\usepackage{hyperref}
\usepackage{url}
\usepackage{xspace}
\usepackage{amsmath}
\usepackage{amssymb}
\usepackage{mathtools}
\usepackage{amsthm}
\usepackage{cancel}
\usepackage{pifont}
\usepackage{afterpage}
\usepackage{algorithm}
\usepackage{algorithmic}
\usepackage[capitalize,noabbrev]{cleveref}
\crefname{appendix}{Appendix}{Appendices}
\Crefname{appendix}{Appendix}{Appendices}

\usepackage{dblfloatfix}

\makeatletter
\DeclareRobustCommand\onedot{\futurelet\@let@token\@onedot}
\def\@onedot{\ifx\@let@token.\else.\null\fi\xspace}
\def\eg{\emph{e.g}\onedot}

\makeatother

\def\ours{\textsc{Best}\xspace}
\def\plus{$^+$\xspace}

\theoremstyle{plain}

\theoremstyle{definition}

\theoremstyle{remark}

\title{Hidden in Plain Sight: The Overlooked Significance of\\Canonical Elements for Extreme LLM Sparsity}

\author{
  Hyeondo Jang \quad Kwanhee Lee \quad Dongyeop Lee \quad Namhoon Lee \\
  POSTECH \\
  \texttt{\{hyeondo.jang, kwanhee.lee, dongyeop.lee2, namhoon.lee\}@postech.ac.kr}
}

\begin{document}

\maketitle

\begin{abstract}

    Large language models (LLMs) are often considered fragile under aggressive sparsification, and maintaining reliable performance typically requires sticking to moderate sparsity levels.
    However, recent studies suggest that LLMs are more resilient to high sparsity than previously thought, reframing the problem as a design challenge rather than a fundamental limitation.
    In this work, we challenge the perceived limits of unstructured post-training LLM pruning by revisiting elementary pruning strategies that have remained relatively underexplored at this scale.
    Through a progressive sparsification framework with second-order saliency and continued training coordinated with sparsity progression, we show that pretrained LLMs can retain strong performance far beyond commonly studied sparsity regimes.
    Across LLaMA-2 and Qwen-3 model families, our approach improves perplexity and downstream accuracy up to 99\% sparsity, surpassing both the current state-of-the-art and representative baselines.
    Precisely, on LLaMA-2-7B, our approach achieves WikiText-2 perplexities of 13.48 and 19.67 at 95\% and 99\% sparsity, respectively, while delivering 3.23$\times$ decoding speedup and 6.21$\times$ memory savings at 95\% sparsity.
    Taken together, our results show that LLMs can be pushed into extreme sparsity while retaining strong performance, providing a foundation for further improving sparse models in this regime.
    
    \end{abstract}

\section{Introduction}

As large language models (LLMs) continue to scale, their deployment incurs substantial compute and memory costs \citep{brown2020language}.
Sparsification, also widely known as pruning, is a critical strategy to mitigate these overheads, offering the potential to drastically reduce model size and accelerate inference \citep{hoefler2021sparsity, agarwalla2024enabling}.
Nevertheless, much of the literature has tended to report sharp degradation in LLM performance beyond moderate sparsity levels around $50$--$70\%$, contributing to a prevailing perception that such degradation may reflect a potential intrinsic limitation of LLMs~\citep{frantar2023sparsegpt, sunsimple}.

\begin{figure}[t]
    \centering
    \includegraphics[width=\columnwidth]{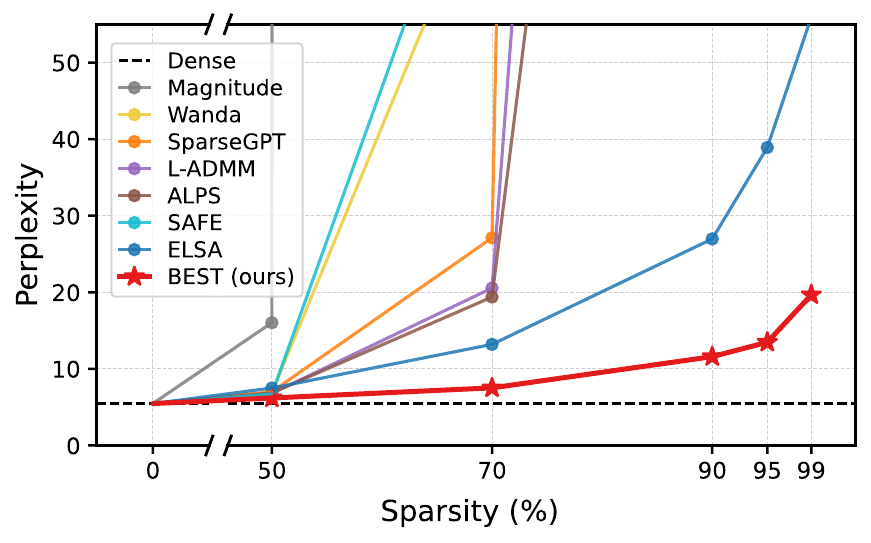}
    \caption{WikiText-2 perplexity of LLaMA-2-7B pruned across a range of sparsity levels.
    We show that extreme sparsity is more realistically attainable than previously suggested.
    See \cref{appendix:comparison-context} for configuration details of the prior pruning approaches shown here.
    }
    \label{fig:elsa_vs_best}
\end{figure}

However, recent work has provided indications that these commonly held preconceptions may warrant reevaluation on multiple fronts \citep{agarwalla2024enabling, lee2025unseen}. 
In particular, emerging evidence suggests that modern LLMs can remain functional well beyond the moderate sparsity regimes often treated as a practical ceiling \citep{agarwalla2024enabling, lee2025unseen}.
Notably, \citet{lee2025unseen} reaches functional performance at 90\% sparsity by performing appropriate optimization through continued training, suggesting that the attainable sparsity of pre-trained LLMs may be higher than previously expected.

This prompts a fundamental inquiry: \emph{to what extent can extreme sparsity be pushed, when paired with a carefully tuned optimization procedure?}
Yet this question has remained largely underexplored: 
the prevailing assumptions that LLMs are too large for retraining or hyperparameter search have favored resource-efficient pruning protocols, which may offer only a partial view of the true sparsity potential in LLMs.

Motivated by this gap, we first revisit the elementary strategies of sparsification that drove neural network pruning: second-order saliency~\citep{lecun1989optimal} and progressive sparsification~\citep{zhu2017prune}, each of which has garnered limited attention in post-training pruning of modern LLMs.
We study how these ingredients should be adapted and scaled for post-training LLM pruning in a continued-training setting.
Surprisingly, we show that, when properly coordinated, these strategies alone can push attainable sparsity in LLMs well beyond previously demonstrated levels.

Across LLaMA-2 and Qwen-3 model families, our approach consistently improves both perplexity and downstream accuracy over prior approaches from $70\%$ to $99\%$ sparsity, with the largest gains in the extreme regime, as previewed on LLaMA-2-7B in \cref{fig:elsa_vs_best} and further demonstrated in \cref{sec:exp}.
Importantly, these gains are not limited to parameter-count reduction: with sparse execution kernels, our pruned LLaMA-2-7B model achieves a $3.23\times$ decoding speedup and a $6.21\times$ reduction in memory footprint over the dense counterpart.

\section{Background}
\label{sec:background}

\subsection{Pruning}

Pruning aims to find sparse neural networks whose weight matrices are mostly populated by zeros, so as to skip corresponding computation and memory allocation while preserving task performance \citep{hoefler2021sparsity}.
A long line of work has demonstrated that pruning can achieve high compression rates with minimal accuracy loss \citep{lecun1989optimal, han2015learning, hoefler2021sparsity}, with resulting sparsity translated into practical speedup through structured formats accelerated on modern GPUs (semi-structured $N{:}M$ sparsity, \eg, $2{:}4$ \citep{mishra2021accelerating, zhou2021learning}, block-structured \citep{gray2017gpu}), or through custom SpMV/SpMM kernels for unstructured patterns \citep{gale2020sparse,macko2025macko}.

Early approaches can largely be characterized by their reliance on a single pruning step (commonly known as one-shot pruning) over the entire training procedure, either at initialization \citep{lee2018snip, tanaka2020pruning, wang2020picking},
or post-hoc on a pretrained model \citep{lecun1989optimal, singh2020woodfisher}.
However, consistently stronger results follow when pruning during training interleaves sparsity with optimization, at the cost of additional compute---either by gradually increasing the sparsity target \citep{zhu2017prune, kurtic2022gmp}, or with allowance for regrowth as in dynamic sparse training \citep{mocanu2018scalable, wortsman2019discovering}.
Among these, iterative pruning \citep{frankle2018lottery, renda2020comparing}, which alternates between pruning and retraining over several cycles, stands out for consistently recovering the strongest sparse subnetworks.

Another key design choice in a pruning strategy is parameter selection: how parameters are chosen for removal.
One widely adopted heuristic scores each parameter by its estimated contribution to the loss, using signals ranging from raw magnitude \citep{han2015learning} to gradient--weight products \citep{lee2018snip} to second-order curvature
\citep{lecun1989optimal, singh2020woodfisher}.
Alternatively, learning-based methods forgo explicit scoring altogether, instead learning the sparsity pattern jointly with the weights via learnable thresholds \citep{kusupati2020soft} or relaxed binary masks.
A separate line of work casts pruning as a constrained optimization problem, with solutions ranging from regularization \citep{louizos2017learning, wen2016learning} to iterative hard thresholding \citep{peste2021ac} and Frank-Wolfe \citep{lu2022learning}.

\subsection{Pruning in LLMs}

With the advent of large language models, however, this design space has largely converged onto a single corner.
Early efforts in LLM pruning focused on computationally inexpensive procedures, driven by the rapid scaling of LLMs that rendered conventional pruning pipelines infeasible.
Post-training, one-shot pruning with layer-wise surrogate objectives \citep{frantar2023sparsegpt, sunsimple} emerged as the dominant approach, demonstrating that moderate sparsity could be obtained with minimal cost and overhead.
These methods established the feasibility of pruning large models, but their reliance on these surrogate objectives imposed a sharp ceiling: they consistently fail to maintain functional performance much beyond roughly $70\%$ sparsity \citep{meng2024alps,huang2025determining}, while the compression levels at which meaningful hardware benefits materialize lie considerably higher~\citep{agarwalla2024enabling, gale2020sparse}.

A growing body of work has sought to push past this ceiling while staying within this paradigm, through better optimization of the layer-wise objective \citep{bovzafast, meng2024alps}, non-uniform sparsity allocation across layers \citep{yin2023outlier, sieberling2024evopress}, or lightweight fine-tuning to compensate for pruning error \citep{zhang2023dynamic, huang2025dynamic}.
These refinements deliver steady but modest gains; none reliably prevent the extreme-sparsity collapse. 
More recent work questions the template itself, broadening the surrogate from per-layer to block-wise or whole-model objectives \mbox{\citep{shin2024rethinking, bai2024sparsellm}}.
Notably, \citet{lee2025unseen} argue that the field's reliance on cost-minimal, one-shot, layer-wise surrogates has been an unduly restrictive self-imposed constraint.
Relaxing this constraint, their iterative, surrogate-free approach demonstrates that functional performance can be retained at $90\%$ sparsity.

Taken together, these results indicate that the prevailing limit on LLM sparsity is not intrinsic, but a consequence of the restricted paradigm through which the field has operated.
Motivated by this, we ask how far extreme sparsity, where practical hardware benefits are substantial, can be pushed when essential pruning components given little attention at modern LLM scale are properly applied.

\section{Method}

\label{sec:method}

Let us start by formulating pruning as a constrained optimization problem:
\begin{equation}
\min_{\mathbf{w}} \mathcal{L}(\mathbf{w}) \quad \text{s.t.} \quad \|\mathbf{w}\|_0 \leq k,
\label{eq:pruning_obj}
\end{equation}
where we minimize the loss $\mathcal{L}:\mathbb{R}^d\to \mathbb{R}$ through optimizing the model parameters $\mathbf{w}\in \mathbb{R}^d$ with $k<d$ number of non-zero weights computed by the $L_0$-norm $\|\cdot\|_0$.

While many strategies exist for solving \cref{eq:pruning_obj}~\citep{zhu2017prune,frankle2018lottery,evci2020rigging,liu2020dynamic}, we build upon the progressive sparsification framework~\citep{zhu2017prune}, a simple yet flexible procedure whose design space exposes the key components governing sparsification performance.

Within this framework, we decompose the pruning process as:
\begin{align}
\mathbf{m}^{t+1} &= \mathcal{M}^t (\mathbf{m}^{t} \odot \mathbf{w}^{t}), \label{eq:mask_compute} \\
\text{s.t.} \, & \quad d = \|\mathbf{m}^{0}\|_0 \geq \dots \geq \|\mathbf{m}^{T}\|_0 = k, \label{eq:sparsity_schedule} \\
\mathbf{w}^{t+1} &= \arg\min_{\mathbf{w}} \mathcal{L}(\mathbf{m}^{t+1} \odot \mathbf{w}), \label{eq:weight_recovery}
\end{align}
where we alternate between the masking operation $\mathcal{M}^t:\mathbb{R}^d\to\{0,1\}^d$ in \cref{eq:mask_compute} (subject to the sparsity schedule in \cref{eq:sparsity_schedule}) and the weight training for the remaining parameters in \cref{eq:weight_recovery}. The process is initialized with the pretrained weights, setting $\mathbf{w}^0 = \mathbf{w}_{\text{pre}}$ with a dense starting mask $\mathbf{m}^0=\mathbf{1}$.

The decomposition above exposes three design components governing sparsification performance:
\begin{itemize}[leftmargin=*]
    \item \textbf{Masking strategy} \eqref{eq:mask_compute} by designing saliency criteria to determine which parameters in $\mathbf{w}$ to remove, along with the sparsity allocation across modules (\cref{subsec:saliency_design});
    \item \textbf{Sparsity schedule} \eqref{eq:sparsity_schedule} by specifying the progression of $\|\mathbf{w}\|_0$ towards the target $k$ (\cref{subsec:schedule_design});
    \item \textbf{Training strategy} \eqref{eq:weight_recovery} by optimizing the remaining active parameters in $\mathbf{w}$ for recovery (\cref{subsec:training_design}).
\end{itemize}
The following subsections present our design choices for each.

\subsection{Masking strategy}
\label{subsec:saliency_design}

The ideal design for the masking function $\mathcal{M}^t$ in Equation \ref{eq:mask_compute} would be one that solves the combinatorial subproblem
\begin{equation}
    \operatorname*{arg\,min}_{\|\mathbf{m}\|_0 \leq k}\, \mathcal{L}(\mathbf{m} \odot \mathbf{w}^{t}).
    \label{eq:masking_subproblem}
\end{equation}
Solving \cref{eq:masking_subproblem} exactly is intractable, so we pursue a tractable saliency-based strategy: a second-order saliency with an optimizer-aware proxy for efficient saliency computation, and global pruning for dynamic sparsity allocation.

\paragraph{Second-order saliency}
A common heuristic for \cref{eq:masking_subproblem} is to design a saliency criterion \citep{hoefler2021sparsity}, assigning a scalar importance score $s_i$ to each parameter $w_i$ based on the expected increase in loss $\mathcal{L}$ upon its removal \citep{han2015learning,lee2018snip,lecun1989optimal}.
Second-order pruning criteria \citep{lecun1989optimal} estimate the loss change induced by removing parameters using a local Taylor expansion:
\begin{equation}
\label{eq:taylor}
    \Delta \mathcal{L}
    \approx
    \mathbf{g}^{\top}\Delta \mathbf{w}
    +
    \frac{1}{2}\Delta \mathbf{w}^{\top} H \Delta \mathbf{w}.
\end{equation}
Since pre-trained models are already converged, the gradient term is expected to be small, i.e., $\mathbf{g}\approx \mathbf{0}$.
Under a diagonal Hessian approximation, this yields the following saliency score for pruning parameter $w_i$:
\begin{equation} 
    s_i^{\mathrm{2nd}}
    =
    \frac{1}{2} H_{ii} w_i^2 .
\label{eq:second_order_saliency}
\end{equation}
This criterion incorporates both parameter magnitude and local loss curvature, providing a direct proxy for the loss increase induced by pruning each parameter.
Despite its appeal, directly using \cref{eq:second_order_saliency} at LLM scale is impractical, as estimating Hessian information for models with billions of parameters is prohibitively expensive.

\begin{table*}[!t]
    \centering
    \scriptsize
    \caption{Perplexities ($\downarrow$) of LLaMA-2 and Qwen-3-Base models pruned to various sparsity levels using different methods. \ours{} consistently outperforms baselines across all settings, with the margin widening at higher sparsity.}
    \label{tab:table1}
    \resizebox{0.85\textwidth}{!}{%
    \begin{tabular}{c l cc cc cc cc}
        \toprule
        & & \multicolumn{4}{c}{LLaMA-2}
          & \multicolumn{4}{c}{Qwen-3} \\
        \cmidrule(lr){3-6} \cmidrule(lr){7-10}
        & & \multicolumn{2}{c}{7B}
          & \multicolumn{2}{c}{13B}
          & \multicolumn{2}{c}{4B-Base}
          & \multicolumn{2}{c}{8B-Base} \\
        \cmidrule(lr){3-4} \cmidrule(lr){5-6} \cmidrule(lr){7-8} \cmidrule(lr){9-10}
        Sparsity & Method
            & Wiki & C4 & Wiki & C4 & Wiki & C4 & Wiki & C4 \\
        \midrule
        0\% & Dense
            & 5.47 & 7.26
            & 4.88 & 6.73
            & 7.82 & 11.58
            & 6.93 & 10.41 \\
        \midrule
        \multirow{7}{*}{70\%}
            & Magnitude         & 4.4e4 & 3.3e4 & 212.6 & 163.7 & 5.8e5 & 5.6e5 & 2.2e6 & 1.1e6 \\
            & Wanda             & 69.91 & 81.88 & 44.41 & 46.24 & 60.77 & 71.14 & 50.17 & 58.61 \\
            & SparseGPT         & 26.02 & 29.85 & 21.28 & 21.30 & 27.95 & 32.93 & 20.43 & 24.74 \\
            & Magnitude\plus{}  & \underline{7.72} & \underline{10.04} & \underline{6.74} & 9.09 & 11.00 & 16.32 & 9.57 & 15.05 \\
            & Wanda\plus{}      & 8.06 & 10.44 & 6.94 & 9.54 & 10.83 & \underline{16.05} & 9.57 & 14.95 \\
            & SparseGPT\plus{}  & 7.74 & 10.07 & 6.84 & \underline{9.09} & \underline{10.68} & 16.92 & \underline{9.27} & \underline{14.62} \\
            \rowcolor{gray!15}\cellcolor{white}
            & \ours{}           & \textbf{7.52} & \textbf{9.56} & \textbf{6.64} & \textbf{8.89} & \textbf{10.15} & \textbf{15.40} & \textbf{9.00} & \textbf{14.06} \\
        \midrule
        \multirow{7}{*}{90\%}
            & Magnitude         & NaN   & NaN   & 8.4e4 & 6.4e4 & 7.9e5 & 8.3e5 & 2.7e6 & 5.7e6 \\
            & Wanda             & 9313  & 7419  & 2.3e4 & 1.4e4 & 1.9e5 & 1.7e5 & 2.8e4 & 1.8e4 \\
            & SparseGPT         & 1514 & 908.4 & 1389 & 797.1 & 1632 & 1319 & 369.1 & 264.8 \\
            & Magnitude\plus{}  & \underline{13.18} & \underline{15.40} & \underline{11.51} & \underline{14.94} & 24.94 & 31.38 & 20.55 & 26.40 \\
            & Wanda\plus{}      & 15.61 & 18.82 & 13.89 & 17.02 & 21.58 & 28.30 & 18.22 & 24.85 \\
            & SparseGPT\plus{}  & 13.41 & 16.04 & 11.79 & 15.23 & \underline{18.95} & \underline{25.01} & \underline{16.14} & \underline{22.76} \\
            \rowcolor{gray!15}\cellcolor{white}
            & \ours{}           & \textbf{11.60} & \textbf{13.71} & \textbf{10.04} & \textbf{12.68} & \textbf{14.90} & \textbf{20.79} & \textbf{13.49} & \textbf{19.18} \\
        \midrule
        \multirow{7}{*}{95\%}
            & Magnitude         & 1.7e5 & NaN   & 5.3e4 & 3.6e4 & 2.2e5 & 2.6e5 & 1.8e6 & 1.8e6 \\
            & Wanda             & 1.4e4 & 1.5e4 & 1.6e5 & 1.2e5 & 1.1e4 & 7076 & 1.1e5 & 9.2e4 \\
            & SparseGPT         & 5619 & 6186 & 3088 & 2174 & 3.3e4 & 2.7e4 & 1857 & 1179 \\
            & Magnitude\plus{}  & 20.52 & 21.50 & 45.48 & 41.05 & 41.11 & 44.81 & 35.63 & 40.69 \\
            & Wanda\plus{}      & 21.87 & 22.47 & 19.14 & 21.73 & \underline{28.86} & \underline{35.88} & 26.42 & 33.52 \\
            & SparseGPT\plus{}  & \underline{18.75} & \underline{20.72} & \underline{16.25} & \underline{19.55} & 29.18 & 36.49 & \underline{23.83} & \underline{30.64} \\
            \rowcolor{gray!15}\cellcolor{white}
            & \ours{}           & \textbf{13.48} & \textbf{16.52} & \textbf{12.72} & \textbf{15.33} & \textbf{17.67} & \textbf{22.86} & \textbf{16.08} & \textbf{22.08} \\
        \midrule
        \multirow{7}{*}{99\%}
            & Magnitude         & NaN   & NaN   & 2.8e4 & 3.1e4 & 2.0e5 & 3.2e5 & 1.1e5 & 1.2e5 \\
            & Wanda             & 6.4e5 & 7.4e5 & 1.5e4 & 2.4e4 & 1.0e5 & 9.9e4 & 3.2e5 & 2.1e5 \\
            & SparseGPT         & NaN  & NaN  & 8466 & 3886 & 4.2e4 & 5.4e4 & 3.8e4 & 4.0e4 \\
            & Magnitude\plus{}  & 50.40 & 45.02 & 67.79 & 57.06 & 77.73 & 71.73 & 50.32 & 52.88 \\
            & Wanda\plus{}      & 44.52 & 38.61 & 35.98 & 35.34 & \underline{60.13} & \underline{64.93} & 46.12 & 51.08 \\
            & SparseGPT\plus{}  & \underline{35.53} & \underline{32.39} & \underline{31.36} & \underline{31.28} & 69.93 & 70.16 & \underline{38.03} & \underline{43.68} \\
            \rowcolor{gray!15}\cellcolor{white}
            & \ours{}           & \textbf{19.67} & \textbf{22.03} & \textbf{22.07} & \textbf{24.15} & \textbf{25.96} & \textbf{31.98} & \textbf{23.39} & \textbf{29.51} \\
        \bottomrule
\end{tabular}}
\end{table*}

\paragraph{Optimizer-aware proxy for efficient saliency computation}
To avoid exact Hessian computation, we approximate $H$ by the Generalized Gauss-Newton matrix \citep{botev2017practical}, which coincides with the Fisher Information Matrix $F$ under the standard cross-entropy objective \citep{martens2020new}.
For converged models, $F$ is commonly approximated by the Empirical Fisher $F = \mathbb{E}[gg^\top]$, whose diagonal entries $F_{ii} = \mathbb{E}[g_i^2]$ provide an efficient curvature proxy.

Even computing $F_{ii}$, however, requires an extra pass to collect and average squared gradients across data samples---a bottleneck at LLM scale.
We instead reuse the running statistics already maintained by the Adam optimizer \citep{kingma2014adam} during retraining: its exponential moving average of squared gradients, $\hat{v}^t$, has been shown to approximate the diagonal Fisher \citep{li2025fishers}.
Substituting $\hat{v}^t$ for $F_{ii}$ in \cref{eq:second_order_saliency} yields a zero-cost saliency score,
\begin{equation}
\hat{s}_i^{\mathrm{2nd}} = \frac{1}{2}\, \hat{v}^t_i\, w_i^2,
\end{equation}
giving curvature-aware pruning at no additional gradient-computation cost.

\paragraph{Global pruning for dynamic sparsity allocation}
Once a saliency criterion is chosen, pruning still requires specifying the group within which scores are compared.
In principle, global comparison --- ranking all parameters jointly under a single threshold --- is the most faithful realization of the constrained objective in \cref{eq:masking_subproblem}, allocating sparsity in proportion to per-parameter importance without any predetermined per-layer ratio.
Prior post-training LLM pruners are nonetheless confined to layer-wise comparison \citep{frantar2023sparsegpt,sunsimple}, a restriction inherited from their layer-wise, local surrogate objectives that offer no signal for cross-layer ranking, forcing uniform per-layer sparsity and motivating separate sparsity-allocation strategies \citep{yin2023outlier,sieberling2024evopress}.
Our second-order saliency, by contrast, directly approximates the global loss change and is therefore comparable across modules, admitting a single global threshold without such strategies.
We accordingly adopt pruning by global comparison.
Implementation-wise, we make this global threshold tractable under distributed training of LLMs via a distributed binary search that keeps every score in place and exchanges only scalar counts, as detailed in \cref{appendix:dbsearch}.

\subsection{Sparsity schedule}
\label{subsec:schedule_design}

The constraint in \cref{eq:sparsity_schedule} only specifies that $\|\mathbf{m}^n\|_0$ must monotonically decrease from $d$ to $k$; the trajectory is a design choice.
Directly projecting $\mathbf{w}_{\mathrm{pre}}$ to the target sparsity can cause a sudden loss of pretrained knowledge, while an overly slow sparsification process may repeatedly interfere with optimization and prevent efficient adaptation. To balance these effects, we employ a cubic gradual sparsity schedule, which is shown effective in prior work \citep
{zhu2017prune, kurtic2022gmp}, and our ablation (\cref{appendix:cubic-vs-linear}) as follows:

\begin{equation}
S_t = S_{\text{final}} + (S_{\text{init}} - S_{\text{final}})\left(1 - \frac{t - t_0}{T_{\text{prune}}}\right)^3,
\label{eq:cubic_schedule}
\end{equation}
where $S_{\text{init}}$ and $S_{\text{final}}$ denote the initial and target sparsity, $t_0$ is the step at which pruning begins, and $T_{\text{prune}}$ is its duration.
In our work, $S_{\text{init}}$ is set to $0$, although other choices are possible.
After step $t_0 + T_{\text{prune}}$ the mask is frozen at $S_{\text{final}}$ and training continues until the total budget is exhausted.

\subsection{Training strategy}
\label{subsec:training_design}

The weight-update step in \cref{eq:weight_recovery} must reconcile two demands of post-training pruning: keeping the early learning rate small to avoid disturbing the near-converged $\mathbf{w}_{\mathrm{pre}}$, and providing a sufficiently large learning rate later to absorb the loss perturbations introduced by the sparsity ramp in \cref{eq:cubic_schedule}.

This leads to a learning rate that starts small, grows with the sparsity ramp, and decays toward convergence---in practice, the standard linear warmup followed by linear decay,
\begin{equation}
\fontsize{10.5}{12.5}\selectfont
\eta_t =
\begin{cases}
\eta_{\mathrm{target}} \cdot \tfrac{t}{T_{\mathrm{warmup}}}, & \text{if } t \leq T_{\mathrm{warmup}}, \\[4pt]
\eta_{\mathrm{target}} \cdot \tfrac{T - t}{T - T_{\mathrm{warmup}}}, & \text{if } t > T_{\mathrm{warmup}},
\end{cases}
\label{eq:lr_schedule}
\end{equation}
where $T_{\mathrm{warmup}}$ is the warmup length and $T$ is the total number of steps.
We empirically find that a short warmup suffices and set $T_{\mathrm{warmup}} = 0.1\,T$ throughout.

Together, these three components form our complete framework: a global second-order saliency with an optimizer-aware proxy for masking, a cubic schedule for sparsity progression, and a learning-rate schedule coordinated with the sparsity progression for weight updates.
We call this framework \ours{}, reflecting our goal of pursuing the best performance by designing these components jointly.
We analyze the design principles behind each choice in \cref{subsec:ablation}.

\FloatBarrier
\begin{figure*}[!t]
    \centering
    \includegraphics[width=\linewidth]{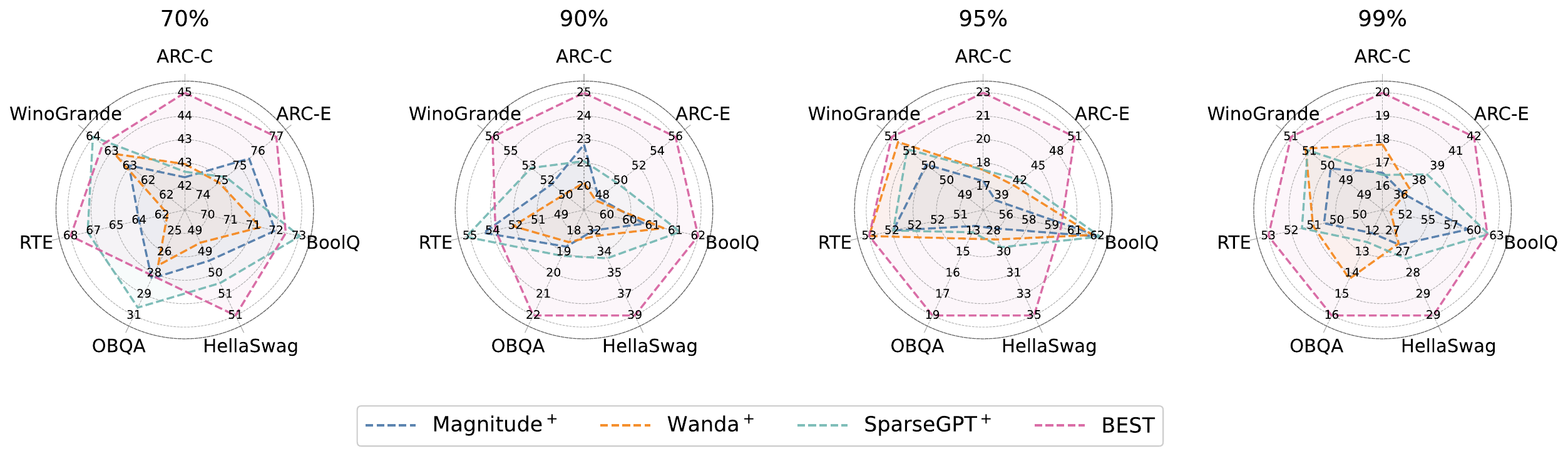}
    \caption{Qwen3-8B zero-shot accuracy across seven tasks at each sparsity. \ours{} outperforms one-shot methods followed by retraining on most tasks. Full numerical results are provided in \cref{appendix:zero-shot}.}
    \label{fig:qwen3_8b_zeroshot_radar}
\end{figure*}

\begin{table}[!t]
    \centering
    \footnotesize
    \begin{minipage}[t]{\linewidth}
        \centering
        \captionof{table}{WikiText-2 perplexity of Qwen2.5-32B. The dense baseline has perplexity $5.12$.}
        \label{tab:qwen32b}
        \vspace{-0.8em}
        \renewcommand{\arraystretch}{1.05}
        \scriptsize
        \setlength{\tabcolsep}{0pt}
        \begin{tabular}{@{}>{\centering\arraybackslash}p{0.20\linewidth}*{4}{>{\centering\arraybackslash}p{0.18\linewidth}}@{}}
            \toprule
            \multirow{2}{*}{Method} & \multicolumn{4}{c}{Sparsity} \\
            \cmidrule(lr){2-5}
            & $70\%$ & $90\%$ & $95\%$ & $99\%$ \\
            \midrule
            SparseGPT\plus{}
              & $7.28$ & $13.93$ & $24.30$ & $37.60$ \\
            \ours{}
              & $\mathbf{6.71}$ & $\mathbf{10.08}$ & $\mathbf{12.98}$ & $\mathbf{21.85}$ \\
            \bottomrule
        \end{tabular}
    \end{minipage}
    \par\vspace{0.8em}
    \begin{minipage}[t]{\linewidth}
        \centering
        \vspace*{0.4em}
        \captionof{table}{Comparison with recent pruning methods on LLaMA-2-7B. Baseline numbers are taken from \citet{lee2025unseen} and therefore serve as contextual comparisons.}
        \label{tab:advanced_methods_comparison}
        \renewcommand{\arraystretch}{1.14}
        \scriptsize
        \setlength{\tabcolsep}{4.2pt}
        \resizebox{0.92\linewidth}{!}{%
        \begin{tabular}{l ccc ccc}
        \toprule
        \multirow{3}{*}{Method}
        & \multicolumn{6}{c}{Sparsity} \\
        \cmidrule(lr){2-7}
        & \multicolumn{3}{c}{70\%}
        & \multicolumn{3}{c}{90\%} \\
        \cmidrule(lr){2-4} \cmidrule(lr){5-7}
        & Wiki & C4 & AZS
        & Wiki & C4 & AZS \\
        \midrule
        L-ADMM            & 20.56 & 22.20 & 44.22 & 400.5 & 287.1 & 31.88 \\
        ALPS              & 19.39 & 20.37 & 45.03 & 248.8 & 180.9 & 32.03 \\
        SAFE              & 86.80 & 48.54 & 38.94 & 1.6e4 & 1.6e4 & 32.63 \\
        SparseLLM         & 37.65 & 35.00 & 40.54 & 1267  & 648.0 & 31.93 \\
        ELSA              & 13.20 & 14.08 & 46.21 & 26.97 & 23.14 & 38.52 \\
        \rowcolor{gray!10}
        \ours{}           & \textbf{7.52} & \textbf{9.56} & \textbf{52.06} & \textbf{11.60} & \textbf{13.71} & \textbf{42.34} \\
        \bottomrule
        \end{tabular}}
    \end{minipage}
\end{table}

\section{Experiments}
\label{sec:exp}

We report our main language-modeling and zero-shot results at extreme sparsity in \cref{subsec:main_results} and comparisons with recent pruning methods in \cref{subsec:further_comparisons}. We then study the practical value of the resulting sparsity in \cref{subsec:value_extreme_sparsity} and analyze the design and cost of \ours{} in \cref{subsec:ablation,subsec:cost}.

\paragraph{Setup}
We evaluate our method on LLaMA-2 \citep{touvron2023llama}, Qwen2.5 \citep{yang2024qwen2}, and Qwen-3 \citep{yang2025qwen3}, using WikiText-2 \citep{merity2017pointer} and C4 \citep{raffel2020exploring} datasets for language modeling evaluation and \texttt{lm-eval-harness} for zero-shot generalization performance.
Detailed setup for training and evaluation can be found at \cref{appendix:setup}.
\paragraph{Baselines}
We compare \ours{} against representative one-shot methods: Magnitude \citep{han2015learning}, Wanda \citep{sunsimple}, and SparseGPT \citep{frantar2023sparsegpt}. To control for training budget, we also evaluate each one-shot method followed by standard retraining, in which the remaining weights are fine-tuned with the mask fixed after pruning, under the same budget as \ours{} (denoted $^+$).
We further compare against recent pruning methods: L-ADMM \citep{bovzafast}, ALPS \mbox{\citep{meng2024alps}}, SAFE \citep{lee2025safe}, SparseLLM \citep{bai2024sparsellm}, and ELSA \citep{lee2025unseen}.

\subsection{Main results}
\label{subsec:main_results}

\paragraph{Language modeling performance}
\label{subsec:language_modeling}

\Cref{tab:table1} reports WikiText-2 and C4 perplexity for LLaMA-2 and Qwen-3 models from $70\%$ to $99\%$ sparsity. \ours{} consistently outperforms all baselines, with the margin widening as sparsity increases.
In the extreme regime ($95\%$--$99\%$), one-shot pruning collapses outright; retraining substantially mitigates this, yet even the strongest baseline (SparseGPT$^+$) reaches only $35.53$ at $99\%$ on LLaMA-2-7B WikiText-2, nearly double \ours{}'s $19.67$.

The same pattern holds across scales and model families: at $99\%$, \ours{} achieves $22.07$ on LLaMA-2-13B and $25.96$ on Qwen3-4B-Base, against $31.36$ and $60.13$ for the respective strongest baselines.
This trend extends to the larger-scale Qwen2.5-32B model, as reported in \cref{tab:qwen32b}, where \ours{} outperforms SparseGPT\plus{} at all sparsity levels.
These results demonstrate that standard retraining pipelines are insufficient for extreme compression, validating the necessity of an optimization procedure specifically designed for this regime.

\begin{figure*}[!t]
    \centering
    \begin{subfigure}[t]{0.48\linewidth}
        \centering
        \includegraphics[width=\linewidth]{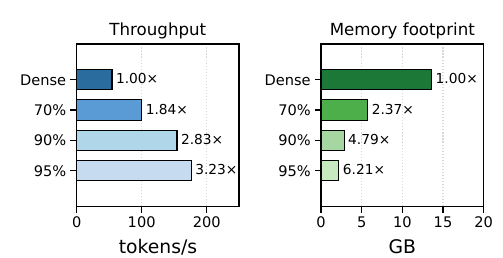}
        \caption{Decoding throughput and memory footprint.}
        \label{fig:inference_efficiency_llama}
    \end{subfigure}
    \hfill
    \begin{subfigure}[t]{0.48\linewidth}
        \centering
        \includegraphics[width=\linewidth]{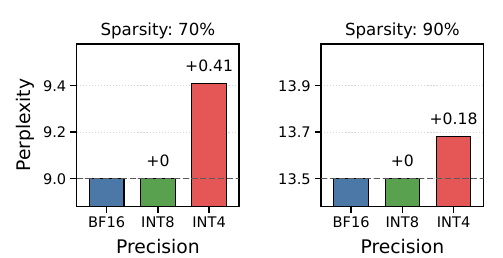}
        \caption{Perplexity change under RTN quantization.}
        \label{fig:sparse_quantized}
    \end{subfigure}
    \caption{(a) End-to-end decoding throughput ($\uparrow$) and memory footprint ($\downarrow$) of LLaMA-2-7B. (b) WikiText-2 perplexity of Qwen-3-8B-Base after RTN quantization (dense BF16: $6.93$); INT8 is lossless, while INT4 causes only minor degradation.}
    \label{fig:compression_efficiency}
\end{figure*}

\paragraph{Zero-shot downstream task performance}
\Cref{fig:qwen3_8b_zeroshot_radar} reports per-task zero-shot accuracy for Qwen3-8B from $70\%$ to $99\%$ sparsity. \ours{} consistently outperforms all baselines, with a clear gap emerging at extreme sparsity.
At $70\%$, methods are competitive: ARC-E $76.35$ vs.\ $74.79$ and HellaSwag $51.17$ vs.\ $50.35$ for SparseGPT$^+$, but SparseGPT$^+$ drops $32.7$ and $21.2$\,\%p on these tasks by $95\%$.
\ours{} retains $50.76$ and $34.91$ at $95\%$ and $42.21$/$29.49$ at $99\%$, outperforming SparseGPT$^+$ on most tasks and on the average at both sparsity levels.
This pattern holds across all four model variants (\cref{appendix:zero-shot}), demonstrating that \ours{} consistently outperforms existing retraining pipelines on downstream tasks across all sparsity levels.

\vspace{-0.5em}

\subsection{Further comparisons with recent pruning methods}
\label{subsec:further_comparisons}
To strengthen our empirical evaluation, we compare \ours{} with recent pruning methods on LLaMA-2-7B in \Cref{tab:advanced_methods_comparison}, including ELSA, which, to the best of our knowledge, established the prior state of the art at extreme sparsity.
\ours{} outperforms all baselines at both $70\%$ and $90\%$ sparsity in perplexity on both WikiText-2 and C4, with the margin widening substantially at $90\%$ sparsity.
At $90\%$ sparsity, \ours{} achieves $11.60$ WikiText-2 perplexity and $13.71$ C4 perplexity, compared with $26.97$ and $23.14$, respectively, for ELSA, the strongest baseline.
The same trend holds for average zero-shot accuracy over the seven downstream tasks evaluated in \cref{subsec:main_results}, with \ours{} achieving $42.34$ versus $38.52$ for ELSA at $90\%$ sparsity.
We extend our evaluation to $95\%$ and $99\%$ sparsity and confirm that \ours{} continues to outperform ELSA, achieving WikiText-2 perplexities of $13.48$ and $19.67$, respectively, compared with $38.91$ and $55.94$ for ELSA, as detailed in \cref{appendix:elsa-comparison}.

\begin{figure*}[!t]
    \centering
    \begin{subfigure}[t]{0.225\linewidth}
        \centering
        \includegraphics[width=\linewidth]{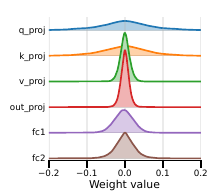}
        \caption{Weight distribution}
        \label{fig:fig4a}
    \end{subfigure}
    \hfill
    \begin{subfigure}[t]{0.73\linewidth}
        \centering
        \includegraphics[width=\linewidth]{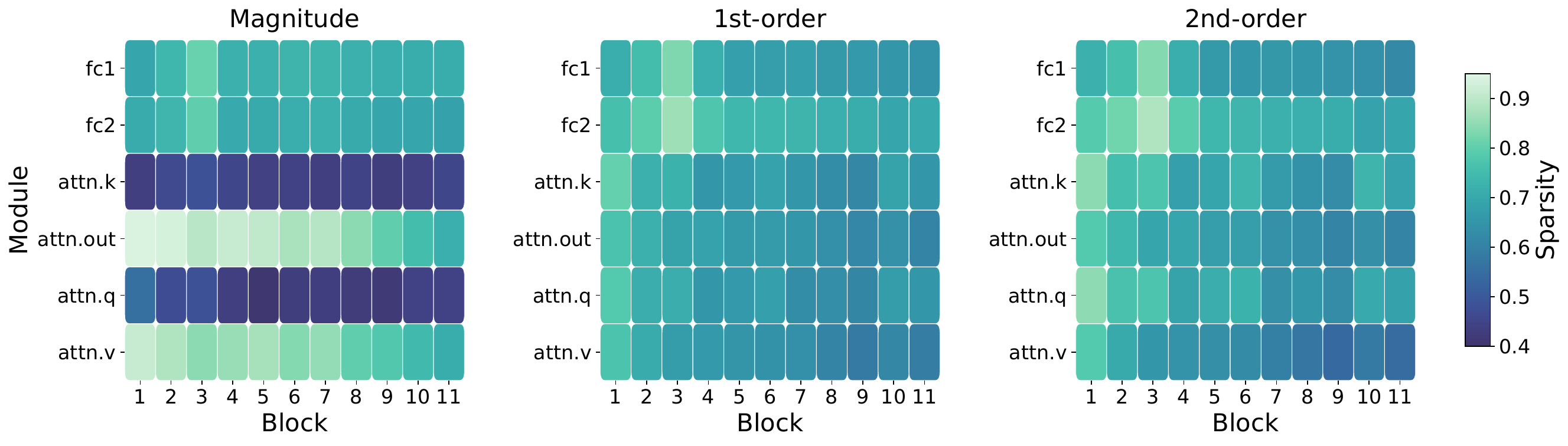}
        \caption{Sparsity allocation heatmaps at 70\% sparsity}
        \label{fig:fig4b}
    \end{subfigure}
    \caption{(a) Per-module weight distributions in OPT-125M first block, shown as a ridgeline with one row per module; different scales make magnitude unreliable as an importance proxy. (b) Per-module sparsity at 70\% across saliency criteria; magnitude assigns excessive sparsity to attention projections, nearly collapsing them.\vspace{-1em}}
    \label{fig:fig4}
\end{figure*}

\gdef\ablationfigurefloat{
\begin{figure*}[!t]
    \centering
    \begin{subfigure}[b]{0.2390\linewidth}
        \centering
        \raisebox{-2.16pt}[\height][\depth]{\includegraphics[width=\linewidth]{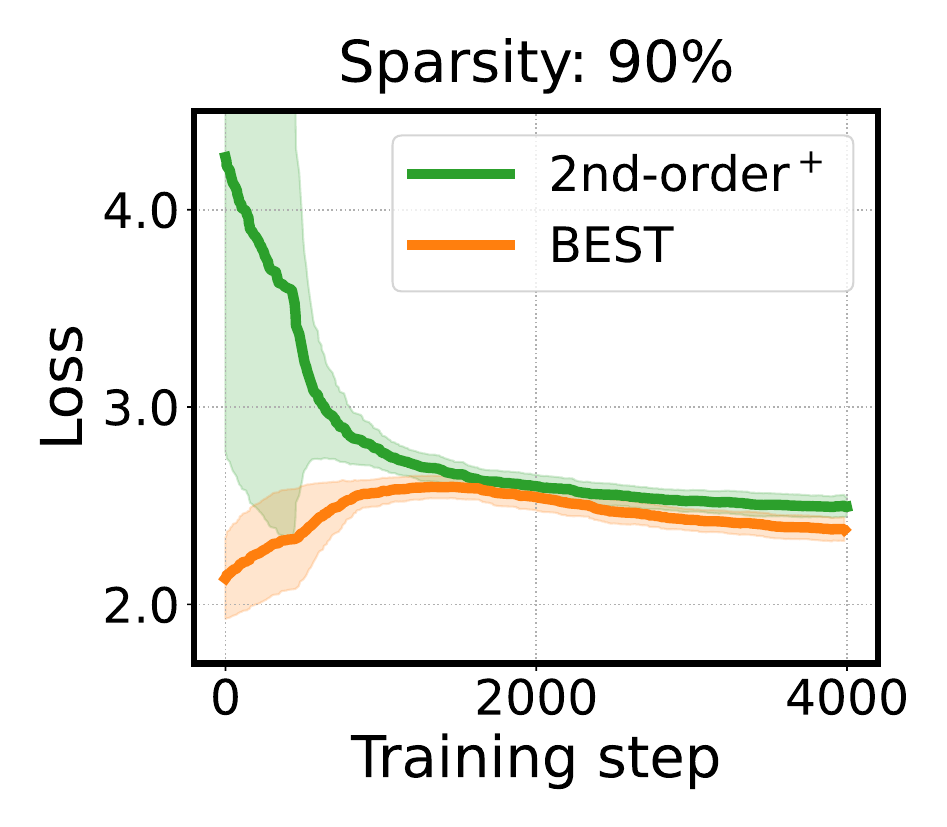}}
        \caption{One-shot vs. Progressive}
        \label{fig:ablation_oneshot_gradual}
    \end{subfigure}
    \hfill
    \begin{subfigure}[b]{0.72\linewidth}
        \centering
        \begin{minipage}[t]{0.32\linewidth}\centering\includegraphics[width=\linewidth]{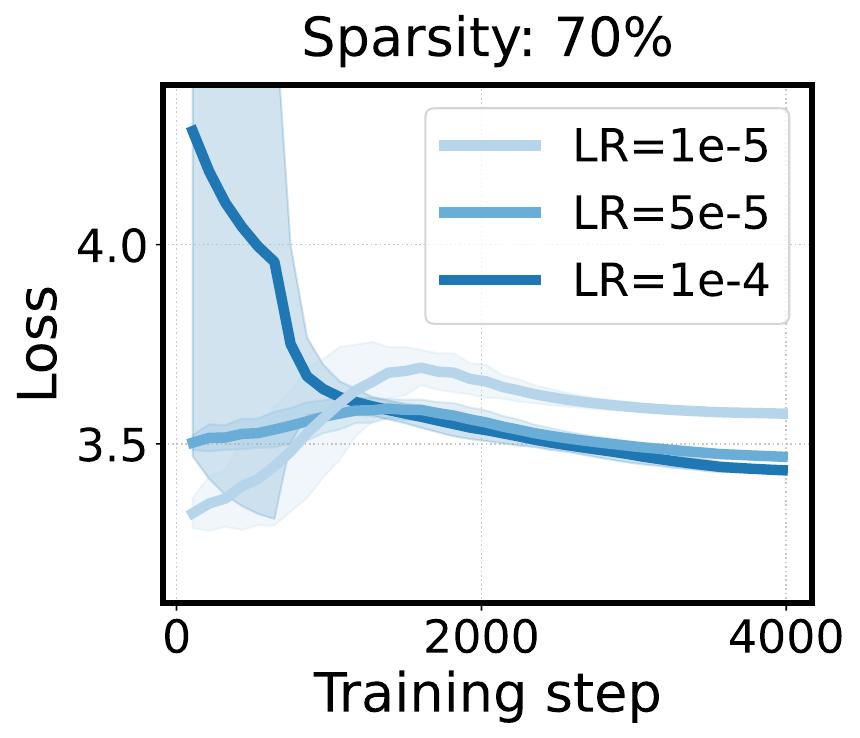}\end{minipage}\hfill
        \begin{minipage}[t]{0.32\linewidth}\centering\includegraphics[width=\linewidth]{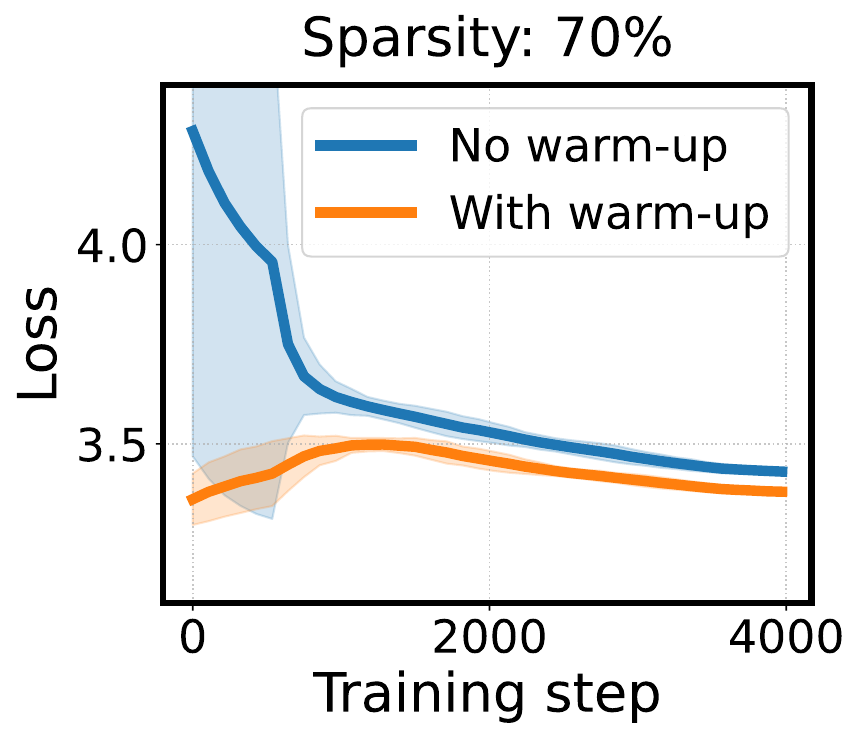}\end{minipage}\hfill
        \begin{minipage}[t]{0.32\linewidth}\centering\includegraphics[width=\linewidth]{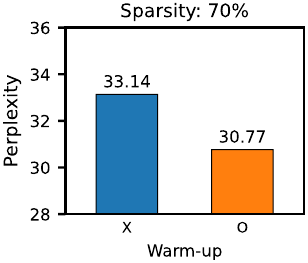}\end{minipage}
        \caption{Learning Rate Warmup}
        \label{fig:ablation_lr_warmup}
    \end{subfigure}
    \caption{(a) Training loss curves for LLaMA-2-7B at $90\%$ sparsity. 2nd-order$^+$ is \ours{} without the gradual schedule: one-shot pruning causes a catastrophic loss increase, while progressive sparsification maintains stable training. (b) Validation loss under different LRs on OPT-125M at $70\%$ sparsity (left): larger LRs are required but applying them directly destabilizes training; a short warm-up resolves the instability by ramping LR gradually (middle), yielding consistently lower WikiText-2 perplexity (right).}
    \label{fig:ablation_combined}
\end{figure*}
}

\subsection{Validating sparsity benefits}
\label{subsec:value_extreme_sparsity}

In this subsection, we assess the practical value of sparsity achieved by \ours{}.
We evaluate whether the resulting sparse models deliver inference efficiency and remain amenable to further compression via quantization.

\paragraph{Inference speedup \& memory savings}
To assess whether parameter reduction leads to actual inference gains, we evaluate our pruned LLaMA-2-7B model in one concrete deployment setting, using the recent SpMV kernel~\citep{macko2025macko} on a single RTX 3090, following its batch-size-$1$, 100-token decode-only protocol.
\cref{fig:inference_efficiency_llama} reports end-to-end decoding throughput and GPU memory footprint including sparse index metadata across sparsity levels.
At $95\%$ sparsity, our model achieves $\mathbf{3.23}\times$ higher decoding throughput than the dense baseline ($176.18$ vs.\ $54.47$ tokens/s) and a memory footprint $\mathbf{6.21}\times$ lower ($2.19$\,GB vs.\ $13.6$\,GB), providing preliminary evidence that extreme sparsity can translate into practical inference gains in this setting.
Notably, this $95\%$-sparse model attains language modeling quality comparable to ELSA's $70\%$-sparse counterpart (WikiText-2 PPL $13.48$ vs.\ $13.20$; \cref{tab:table1,tab:advanced_methods_comparison}). In other words, \ours{} realizes the above speedup and memory gains while preserving the language modeling quality of prior sparse models.\vspace{-0.1em}

\paragraph{Pruning and quantization}

Stable sparse models can also be further compressed through quantization.
We apply simple round-to-nearest (RTN) post-training quantization to the remaining nonzero weights of \ours{}-pruned Qwen-3-8B-Base.
As shown in \cref{fig:sparse_quantized}, INT8 quantization is almost lossless at either sparsity level,
while INT4 incurs a modest increase at $70\%$ ($+0.41$), and an even smaller degradation at $90\%$ ($+0.18$).
These results suggest that the sparse models produced by \ours{} remain stable under low-precision compression, providing a simple path to combine pruning and quantization for further compression.\vspace{-0.2em}

\afterpage{%
\begin{table}[!t]
    \centering
    \footnotesize
    \caption{Effect of saliency and comparison group on OPT-125M perplexity; 2nd-order consistently achieves the best, and global pruning further improves it.}
    \label{tab:comparison_group}
    \renewcommand{\arraystretch}{1.05}
    \fontsize{7.5pt}{9pt}\selectfont
    \setlength{\tabcolsep}{4.2pt}
    \begin{tabular*}{0.92\linewidth}{@{\extracolsep{\fill}}c c cc cc@{}}
        \toprule
        &
        & \multicolumn{4}{c}{Sparsity} \\
        \cmidrule(lr){3-6}
        &
        & \multicolumn{2}{c}{70\%}
        & \multicolumn{2}{c}{90\%} \\
        Saliency & Group
        & Wiki & C4
        & Wiki & C4 \\
        \midrule
        \multirow{2}{*}{Magnitude}
        & Layer
        & \textbf{32.44} & \textbf{31.29}
        & \textbf{47.62} & \textbf{42.28}  \\
        & Global
        & 34.54 & 31.41
        & 54.79 & 46.43 \\
        \midrule
        \multirow{2}{*}{1st-order}
        & Layer
        & 36.61 & 34.85
        & 56.32 & 48.07 \\
        & Global
        & \textbf{36.49} & \textbf{33.00}
        & \textbf{54.30} & \textbf{45.50} \\
        \midrule
        \multirow{2}{*}{2nd-order}
        & Layer
        & 31.28 & 30.84
        & 44.49 & 41.23 \\
        & Global
        & \textbf{30.77} & \textbf{29.16}
        & \textbf{43.84} & \textbf{39.15} \\
        \bottomrule
    \end{tabular*}

\end{table}
}

\subsection{Ablation study}
\label{subsec:ablation}

In this section, to assess the contribution of each to extreme-sparsity performance, we conduct an ablation study on the three components of \ours{}:
masking strategy (\cref{subsec:saliency_design}), progressive sparsity schedule (\cref{subsec:schedule_design}), and training strategy (\cref{subsec:training_design}).
All variants share the same training and evaluation setup; full details and additional sensitivity analyses are in \cref{appendix:ablation}.\vspace{-0.1em}

\paragraph{Masking strategy}

We compare three saliency criteria: magnitude ($|w_i|$), which assigns importance by parameter scale alone; first-order, which captures sensitivity via the gradient term from \cref{eq:taylor} ($|g_i w_i|$); and second-order, which additionally accounts for curvature. 
Following the same zero-cost principle, both gradient-based criteria substitute Adam optimizer states for raw gradients: $|\hat{m}^t_i w_i|$ for first-order and $\tfrac{1}{2}\hat{v}^t_i w_i^2$ for second-order (see \cref{appendix:statistics} for additional ablation on using optimizer states).
Each criterion is evaluated under both layer-wise and global comparison.

\afterpage{\ablationfigurefloat}

\Cref{tab:comparison_group} shows that global second-order pruning achieves the best configuration across all sparsity levels.
Notably, second-order saliency improves further under global comparison ($-0.65$ at $90\%$ Wiki), indicating that its scores are reliable across modules, whereas magnitude degrades under the same setting ($+7.17$ at 90\% Wiki).
To understand this gap, we examine the behavior of magnitude saliency under global sparsity allocation.
As shown in \cref{fig:fig4} (a), weight distributions vary substantially across modules, causing magnitude-based global allocation to be biased by module scale rather than functional importance.
This bias is reflected in \cref{fig:fig4} (b), where magnitude saliency disproportionately allocates sparsity: it over-prunes (\texttt{attn.out, attn.v}), nearly collapsing them even at $70\%$ global sparsity, while leaving (\texttt{attn.q, attn.k}) relatively dense.
By contrast, first- and second-order criteria yield more balanced allocations, with second-order saliency further improving performance by incorporating curvature.
Additional analysis is provided in \cref{appendix:sparsity-allocation}.\vspace{-0.2em}

\paragraph{Sparsity schedule}
To validate the necessity of progressive sparsification, we compare \ours{} against 2nd-order\plus{}, which removes only the gradual sparsity schedule by performing a single pruning step using \ours{}'s saliency before retraining.
As shown in \cref{fig:ablation_oneshot_gradual}, one-shot pruning imposes the target sparsity abruptly, driving the model into a high-loss regime early in optimization and leading to suboptimal solutions.
In contrast, our progressive sparsification avoids this failure mode by introducing sparsity incrementally and maintaining stable optimization, yielding a final loss of $2.38$ and perplexity of $11.60$ at $90\%$ sparsity, compared with $2.49$ and $14.11$ for 2nd-order\plus{}, respectively.
These results support the necessity of a progressive schedule to reach extreme sparsity reliably.
Additional results across sparsity levels, including the resulting perplexities and a comparison with a competitive retrained one-shot baseline, are provided in \cref{appendix:no-schedule}.\vspace{-0.2em}

\paragraph{Training strategy}

With progressive sparsification in place, we next ablate the training strategy.
\cref{fig:ablation_lr_warmup} (left) shows that a larger learning rate is needed for better convergence under high sparsity.
However, applying this large learning rate immediately to pretrained models causes an initial loss spike (\cref{fig:ablation_lr_warmup}, middle), whereas a short warm-up mitigates the instability and yields lower perplexity (\cref{fig:ablation_lr_warmup}, right).
The same improvement carries over to $90\%$ sparsity; additional results and warm-up-length ablations are in \cref{appendix:warmup-choice}.

\subsection{Cost analysis and budget-matched comparisons}
\label{subsec:cost}

\afterpage{%
\begin{table}[!t]
    \centering
    \footnotesize
    \caption{Cost of pruning LLaMA-2-7B to $95\%$ sparsity on A100 80GB GPUs and the resulting WikiText-2 perplexity for each run.}
    \label{tab:cost}
    \renewcommand{\arraystretch}{1.0}
    \fontsize{7.5pt}{9pt}\selectfont
    \setlength{\tabcolsep}{5.4pt}
    \begin{tabular*}{\linewidth}{@{\extracolsep{\fill}}l r r r r@{}}
        \toprule
        Method & Training tokens & \#GPUs & Wall-clock (h) & PPL \\
        \midrule
        SparseGPT           & $0$    & $1$ & $0.25$ & $5619$ \\
        \midrule
        ELSA                & $67$M  & $4$ & $1.87$ & $38.91$ \\
        \ours{}$^{\ddagger}$ & $67$M  & $4$ & $1.74$ & $\mathbf{24.34}$ \\
        \midrule
        SparseGPT\plus{}    & $524$M & $4$ & $10.4$ & $18.75$ \\
        \ours{}             & $524$M & $4$ & $10.3$ & $\mathbf{13.48}$ \\
        \bottomrule
    \end{tabular*}
\end{table}
}

\Cref{tab:cost} reports the pruning cost and resulting perplexity at $95\%$ sparsity.
SparseGPT, a representative one-shot method, is highly cost-efficient, completing in only $0.25$ hours on a single GPU.
However, its perplexity of $5619$ highlights the limitation of cheap one-shot pruning in this extreme regime, motivating additional optimization and shifting the relevant comparison to model quality at comparable training cost.

To examine this trade-off under continued training, we first compare \ours{} with ELSA \citep{lee2025unseen}.
ELSA established the prior state of the art at extreme sparsity by directly optimizing the original language-modeling objective via ADMM \citep{boyd2011distributed}.
ELSA uses C4 with a $67$M-token budget, whereas our main setting uses SlimPajama with a $524$M-token budget because \ours{} remains substantially undertrained under ELSA's setting.
For a controlled cost analysis, we run \ours{} using the same dataset and token budget as ELSA and denote this matched variant as \ours{}$^{\ddagger}$.
\ours{}$^{\ddagger}$ achieves lower perplexity than ELSA ($24.34$ vs.\ $38.91$) with slightly lower wall-clock time ($1.74$ vs.\ $1.87$ hours).
This advantage persists at other sparsity levels, as detailed in \cref{appendix:elsa-comparison}.
Turning to our main $524$M-token setting, \ours{} also outperforms SparseGPT\plus{} at comparable wall-clock time ($13.48$ vs.\ $18.75$ perplexity; $10.3$ vs.\ $10.4$ hours).

Overall, under matched training budgets, \ours{} achieves strong performance without additional wall-clock overhead, demonstrating a favorable performance--cost trade-off enabled by an efficient implementation that reuses optimizer states for saliency estimation and employs distributed binary search for global thresholding.

\section{Conclusion}
\label{sec:conclusion}

Extreme sparsity in LLMs is an increasingly important problem, and a growing body of work has sought to push the limits of how sparse a model can be while retaining its capabilities.
In this work, we find that faithfully applying essential components of the pruning process proves sufficient to establish a new state of the art, with the resulting sparse models yielding measurable inference speedup on commodity hardware.
We hope our framework and analysis provide a foundation for further innovations on extreme LLM sparsity, an active research target.

\section*{Limitations}
\label{sec:limitations}

Several limitations remain in this work, which we plan to address in future research.

First, although \ours{} achieves strong performance at extreme sparsity, its optimization procedure requires continued training and thus incurs additional computational cost. Reducing this cost remains important, and selecting training data so that fewer tokens suffice \citep{lin2024not,xia2024sheared} is one promising direction.

Second, despite the minimal increase in perplexity, the decline in downstream task performance may limit the practical utility of these compressed models. This limitation becomes more pronounced in free-form generation and reasoning: as reported in \cref{appendix:generation}, performance on GSM8K and NQ-Open deteriorates, particularly in the extreme-sparsity regime.
Our pipeline operates on a base pretrained model and therefore defers downstream performance recovery to solution-quality-oriented techniques such as post-training alignment \citep{touvron2023llama,yang2025qwen3}.
Exploring the synergy between extreme sparsification and modern alignment pipelines such as GRPO \citep{shao2024deepseekmath}, together with targeted data curation \citep{penedo2023refinedweb,li2024datacomp}, is a natural direction for recovering nuanced reasoning capabilities.

Moreover, our experiments focus on dense Transformers. Whether the same design principles extend to other architectures, such as Mixture-of-Experts models \citep{jiang2024mixtral}, is an important question that we leave to future work.

Translating strong performance at extreme sparsity into practical inference gains remains contingent on a dedicated sparse kernel \citep{macko2025macko} in the absence of native hardware support for irregular sparsity \citep{agarwalla2024enabling}. This dependence exposes a deployment trade-off: structured and semi-structured methods adopt hardware-friendly patterns that make acceleration more readily attainable on commodity hardware \citep{mishra2021accelerating,zhou2021learning}, albeit with greater difficulty preserving model quality under stronger structural constraints. More efficient kernels and broader hardware support are therefore important for fully realizing the value of extreme sparsity.

Finally, a practically relevant comparison is against small dense models at comparable effective parameter counts.
We outperform such counterparts even with our static framework (\cref{appendix:small-dense-vs-large-sparse}), but the margin remains narrow; incorporating dynamic sparsity mechanisms \citep{evci2020rigging,liu2020dynamic} or knowledge distillation \citep{muralidharan2024compact} may be one way to widen it.

\section*{Acknowledgments}

This work was partly supported by
the Institute of Information \& communications Technology Planning \& Evaluation (IITP)
grant funded by the Korean government (MSIT)
(RS-2019-II191906, Artificial Intelligence Graduate School Program (POSTECH);
RS-2024-00457882, National AI Research Lab Project;
RS-2026-25507427, Development of Efficient Architectures and Training Techniques
for High-Performance Lightweight AI Models),
the Korea Basic Science Institute (National Research Facilities and Equipment Center)
grant funded by the Korean government (MSIT) (RS-2026-25500419)
and the National Research Foundation of Korea (NRF)
grant funded by the Korean government (MSIT)
(RS-2023-00210466,
RS-2025-02264052).
Hyeondo Jang was supported by Kwanjeong Educational Foundation Scholarship.

\bibliography{reference}

\newpage

\begin{appendix}
    \crefalias{section}{appendix}
    \crefalias{subsection}{appendix}
    \section{Additional Discussion}
    \subsection{Further details on prior approaches}
    \label{appendix:comparison-context}

    Here, we provide a more detailed description of the pruning procedures adopted in prior work.
    Magnitude prunes weights directly by their magnitude, whereas Wanda \citep{sunsimple} and SparseGPT \citep{frantar2023sparsegpt} are one-shot methods that determine pruning masks using information from a small calibration dataset.
    ALPS \citep{meng2024alps}, L-ADMM \citep{bovzafast}, and SAFE \citep{lee2025safe} instead perform iterative layer-wise optimization using calibration data.
    ELSA \citep{lee2025unseen} directly optimizes the original language-modeling objective on additional data in a continued-training setting, but we find that this setting leaves \ours{} undertrained and therefore adopt a different one for our main results.
    Budget-matched comparisons with one-shot methods and ELSA are provided in \cref{subsec:main_results,subsec:cost}, respectively, with \ours{} achieving better performance in both cases.

    \section{Experimental details}
    \label{appendix:experimental-details}
    
    \subsection{Implementation and reproducibility details}
    \label{appendix:implementation}
    Our experiments are implemented in PyTorch~\citep{paszke2019pytorch}, with model architectures and datasets managed through the HuggingFace \texttt{transformers} and \texttt{datasets} libraries.
    Our training pipeline is implemented on top of the HuggingFace \texttt{Trainer} framework and supports distributed execution via PyTorch FSDP-2~\citep{zhao2023pytorch}, integrated with HuggingFace Accelerate.
    Unless otherwise noted, all experiments are conducted using a single unified codebase.
    The code is available at \url{https://github.com/LOG-postech/hidden-in-plain-sight}.

    \paragraph{Hardware}
    Experiments are executed on NVIDIA A6000, A100, PRO6000, and H200 GPUs.
    Computational resources are scaled with model size: LLaMA-2-7B/13B and Qwen3-4B/8B are trained using 2--4 NVIDIA A100 GPUs (or PRO6000 GPUs) depending on the experimental configuration, while smaller models such as OPT-125M are trained on a single A6000 GPU.
    For Qwen2.5-32B, we use 8 PRO6000 GPUs and H200 GPUs.

    \paragraph{Efficient global thresholding in a distributed setup}
    \label{appendix:dbsearch}
    Global pruning requires the threshold $\tau$ to be the global $(N-k)$-th smallest among the $N$ saliency scores, where $k$ is the target number of nonzero weights.
    The naive realization gathers the $N$ sharded scores onto a single device and selects the $(N-k)$-th element there, costing $\mathcal{O}(N)$ memory on that device and $\mathcal{O}(N)$ communication; the gather alone becomes a non-trivial bottleneck at LLM scale.
    We instead obtain the same threshold without any gather, via the distributed binary search of \cref{alg:dbsearch}: each of the $G$ ranks (i.e., devices) keeps its score shard in place, and only scalar counts are exchanged.
    Since our saliency criteria are non-negative, we fix $\tau_{\min}{=}0$ and obtain $\tau_{\max}$ via a single \textsc{AllReduce-Max}; the bracket is then halved $T{=}50$ times, with each iteration requiring one scalar \textsc{AllReduce-Sum} of the per-rank counts $c_r$.
    Both the $\mathcal{O}(N)$ memory and $\mathcal{O}(N)$ communication costs of the gather are eliminated, replaced by only $T{+}1$ scalar reductions and a per-rank transient boolean mask of size $\mathcal{O}(N/G)$---only one byte per parameter.
    This yields the global $(N-k)$-th-value threshold after just $T$ halvings.

    \begin{algorithm}[t]
    \caption{Distributed binary search for global threshold $\tau$.}
    \label{alg:dbsearch}
    \begin{algorithmic}[1]
        \REQUIRE Saliency scores $\{s_i\}_{i=1}^{N}$ partitioned across $G$ ranks, with $I_r$ denoting the indices held on rank $r$; target nonzero count $k$; iterations $T$
        \STATE $\tau_{\min} \gets 0$
        \STATE $\tau_{\max} \gets \textsc{AllReduce-Max}\bigl( \max_{i \in I_r} s_i \bigr)$
        \FOR{$t = 1, \dots, T$}
            \STATE $\tau \gets (\tau_{\min} + \tau_{\max}) / 2$
            \STATE $c_r \gets |\{\, i \in I_r : s_i \le \tau \,\}|$
            \STATE $C \gets \textsc{AllReduce-Sum}(c_r)$
            \IF{$C \ge N-k$}
                \STATE $\tau_{\max} \gets \tau$
            \ELSE
                \STATE $\tau_{\min} \gets \tau$
            \ENDIF
        \ENDFOR
        \STATE \textbf{return} $\tau_{\max}$
    \end{algorithmic}
    \end{algorithm}
    \subsection{Detailed experimental setup}
    \label{appendix:setup}
    \paragraph{Calibration data}
    To perform one-shot pruning for Wanda$^+$ and SparseGPT$^+$, we follow the standard one-shot pruning protocol of \citet{frantar2023sparsegpt}.
    Specifically, we sample 128 calibration sequences from the C4 dataset with a sequence length of 2048, which are used solely for one-shot pruning.
    \paragraph{Training data and details.}
    Except for \ours{}$^{\ddagger}$, which uses C4 to match ELSA's setting, all continued-training experiments use SlimPajama\footnote{\href{https://huggingface.co/datasets/DKYoon/SlimPajama-6B}{DKYoon/SlimPajama-6B}}, a cleaned and deduplicated version of RedPajama that combines pretraining corpora from diverse sources including C4, ArXiv, and GitHub.

    For both \ours{} and the one-shot+retrain baselines in our main setting, training data are constructed by randomly sampling non-overlapping token sequences of length 2{,}048.
    To ensure a controlled comparison, the total training budget is fixed at $64 \times 4000 \times 2048$ tokens (batch size 64, training steps 4{,}000, and sequence length 2{,}048).
    For ELSA and \ours{}$^{\ddagger}$, we instead use ELSA's setting of batch size 8, 4{,}096 training steps, and sequence length 2{,}048, corresponding to approximately 67M training tokens.

    Model parameters and optimizer states are kept in full precision, while forward/backward passes use bf16 automatic mixed precision for efficiency.

    \paragraph{Progressive sparsification}
    For \ours{}, given a total training budget of $T_{\text{total}}$ steps, we set the pruning phase to $T_{\text{prune}}=0.5\,T_{\text{total}}$, start from $S_{\text{init}}=0$, and perform $K=50$ mask updates uniformly within this phase, so that masks are updated every $\Delta t = T_{\text{prune}}/K$ steps (40 steps when $T_{\text{total}}=4000$).
    At each mask update, the cubic schedule in \cref{eq:cubic_schedule} determines the new sparsity level, the global mask is recomputed from the current weights, and gradient updates resume on the remaining parameters.
    After $T_{\text{prune}}$, the mask is frozen at the final target sparsity and the remaining steps are used to train the fixed sparse model.
    A full list of hyperparameter configurations is provided in \cref{tab:global_hparams,tab:lr_ours,tab:lr_oneshot}.
    
    \makeatletter
    \setlength{\@dblfpsep}{14pt}
    \makeatother

    \begin{table*}[!t]
    \centering
    \caption{Global hyperparameters of \ours{} and one-shot+retrain strategy shared across all models.}
    \label{tab:global_hparams}
    \small
    \begin{tabular}{l c c}
    \toprule
    Hyperparameter & \ours{} & one-shot+retrain \\
    \midrule
    LR schedule        & Linear decay & Linear decay \\
    Pruning interval $k$ & 40 & -- \\
    Adam $(\beta_1, \beta_2)$ & $(0.9, 0.999)$ & $(0.9, 0.999)$ \\
    Batch size         & 64 & 64 \\
    Training steps     & 4000 & 4000 \\
    Warm-up steps     & 400 & 400 \\
    \bottomrule
    \end{tabular}
    \end{table*}

    \begin{table*}[!t]
    \centering
    \caption{Learning rate ($\eta$) configurations of \ours{} across sparsity levels.}
    \label{tab:lr_ours}
    \small
    \begin{tabular}{c c c c c c}
    \toprule
    Sparsity & OPT-125M & LLaMA-2-7B & LLaMA-2-13B & Qwen3-4B-Base & Qwen3-8B-Base \\
    \midrule
    
    70\%
    & \cellcolor{gray!20}$5\mathrm{e}{-4}$
    & \cellcolor{gray!10}$5\mathrm{e}{-5}$
    & \cellcolor{gray!10}{$5\mathrm{e}{-5}$}
    & \cellcolor{gray!10}$5\mathrm{e}{-5}$
    & \cellcolor{gray!15}$1\mathrm{e}{-4}$ \\

    90\%
    & \cellcolor{gray!25}$1\mathrm{e}{-3}$
    & \cellcolor{gray!15}$1\mathrm{e}{-4}$
    & \cellcolor{gray!15}{$1\mathrm{e}{-4}$}
    & \cellcolor{gray!20}$5\mathrm{e}{-4}$
    & \cellcolor{gray!20}$5\mathrm{e}{-4}$ \\
    
    95\%
    & --
    & \cellcolor{gray!20}$5\mathrm{e}{-4}$
    & \cellcolor{gray!15}{$1\mathrm{e}{-4}$}
    & \cellcolor{gray!20}$5\mathrm{e}{-4}$
    & \cellcolor{gray!20}$5\mathrm{e}{-4}$ \\
    
    99\%
    & --
    & \cellcolor{gray!20}$5\mathrm{e}{-4}$
    & \cellcolor{gray!15}{$1\mathrm{e}{-4}$}
    & \cellcolor{gray!20}$5\mathrm{e}{-4}$
    & \cellcolor{gray!25}$1\mathrm{e}{-3}$ \\

    \bottomrule
    \end{tabular}
    \end{table*}

    \begin{table*}[!t]
    \centering
    \caption{Learning rate ($\eta$) configurations of one-shot+retrain strategy across sparsity levels.}
    \label{tab:lr_oneshot}
    \small
    \begin{tabular}{c c c c c c}
    \toprule
    Sparsity & OPT-125M & LLaMA-2-7B & LLaMA-2-13B & Qwen3-4B-Base & Qwen3-8B-Base \\
    \midrule
    
    70\%
    & \cellcolor{gray!20}$5\mathrm{e}{-4}$
    & \cellcolor{gray!15}$1\mathrm{e}{-4}$
    & \cellcolor{gray!10}{$5\mathrm{e}{-5}$}
    & \cellcolor{gray!15}$1\mathrm{e}{-4}$
    & \cellcolor{gray!15}$1\mathrm{e}{-4}$ \\
    
    90\%
    & \cellcolor{gray!25}$1\mathrm{e}{-3}$
    & \cellcolor{gray!15}$1\mathrm{e}{-4}$
    & \cellcolor{gray!15}{$1\mathrm{e}{-4}$}
    & \cellcolor{gray!20}$5\mathrm{e}{-4}$
    & \cellcolor{gray!20}$5\mathrm{e}{-4}$ \\
    
    95\%
    & --
    & \cellcolor{gray!20}$5\mathrm{e}{-4}$
    & \cellcolor{gray!15}{$1\mathrm{e}{-4}$}
    & \cellcolor{gray!20}$5\mathrm{e}{-4}$
    & \cellcolor{gray!20}$5\mathrm{e}{-4}$ \\
    
    99\%
    & --
    & \cellcolor{gray!20}$5\mathrm{e}{-4}$
    & \cellcolor{gray!15}{$1\mathrm{e}{-4}$}
    & \cellcolor{gray!20}$5\mathrm{e}{-4}$
    & \cellcolor{gray!25}$1\mathrm{e}{-3}$ \\

    \bottomrule
    \end{tabular}
    \end{table*}

    \paragraph{Evaluation}
    We evaluate language-modeling perplexity on the held-out validation splits of Wikitext-2 (\texttt{Salesforce/wikitext-2-raw-v1}) \citep{merity2017pointer} and C4 (\texttt{allenai/c4}, English subset) \citep{raffel2020exploring}, following the standard perplexity protocol established by \citet{frantar2023sparsegpt} and adopted in subsequent LLM pruning work \citep{sunsimple,meng2024alps}: each model uses its own pretrained tokenizer (HuggingFace \texttt{AutoTokenizer}) with a fixed context length of $L=2048$, the validation text is tokenized once and split into contiguous, non-overlapping chunks of length $L$ (no sliding-window stride), and each chunk is fed to the model as a single sequence to compute token-level cross-entropy with the standard shift-by-one labelling.
    We evaluate the entire Wikitext-2 validation split and the first $256\times L$ tokens of the C4 validation split, applying the same protocol to both dense and sparse models.
    For zero-shot accuracy we use \texttt{lm-eval-harness} on seven standard benchmarks --- ARC-Easy and ARC-Challenge (ARC-E/C) \citep{clark2018think}, BoolQ \citep{clark2019boolq}, HellaSwag \citep{zellers2019hellaswag}, OpenBookQA (OBQA) \citep{mihaylov2018can}, RTE \mbox{\citep{wang2018glue}}, and Winogrande \citep{sakaguchi2021winogrande} --- and report the mean accuracy across the seven tasks.

    \clearpage
    \makeatletter
    \setlength{\@dblfpsep}{8pt plus 2fil}
    \makeatother

    \newcounter{figurebeforeappendixb}
    \setcounter{figurebeforeappendixb}{\value{figure}}
    \newcommand{\warmupfiguretop}{%
        \begin{@twocolumnfalse}
        \captionsetup{type=figure}
        \centering
        \begin{subfigure}[t]{0.48\textwidth}
            \centering
            \includegraphics[width=0.49\linewidth]{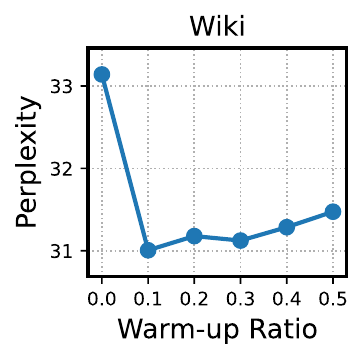}%
            \includegraphics[width=0.49\linewidth]{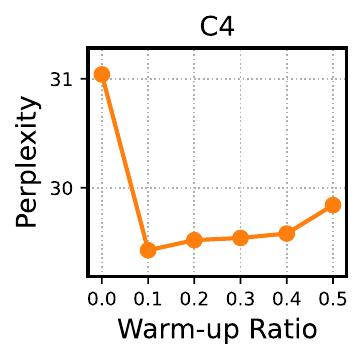}
            \caption{Sparsity: 70\%}
        \end{subfigure}
        \begin{subfigure}[t]{0.48\textwidth}
            \centering
            \includegraphics[width=0.49\linewidth]{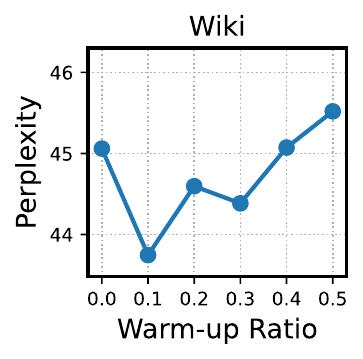}%
            \includegraphics[width=0.49\linewidth]{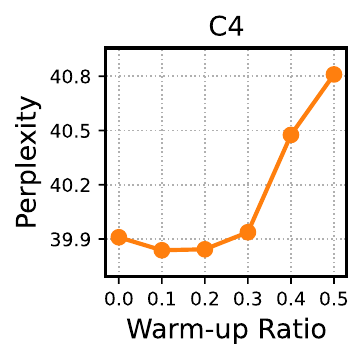}
            \caption{Sparsity: 90\%}
        \end{subfigure}
        \setcounter{figure}{\value{figurebeforeappendixb}}
        \captionof{figure}{Effect of warm-up ratio (of total 4000 steps) on perplexity for OPT-125M across sparsity levels; a short warm-up ($\sim$10\%) consistently achieves stable performance.}
        \label{fig:warmup_choice}

        \vspace{8pt}
        \begin{subfigure}[t]{0.27\textwidth}
            \centering
            \includegraphics[width=\linewidth]{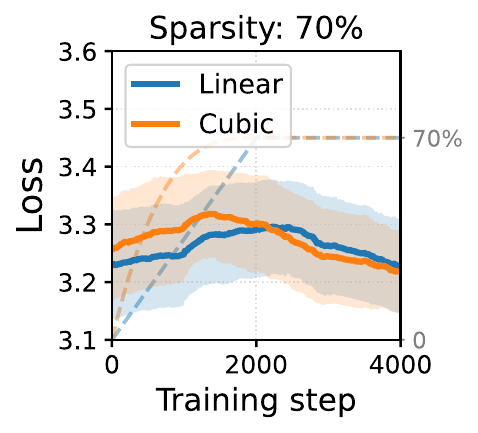}
        \end{subfigure}
        \hspace{0.04\textwidth}
        \begin{subfigure}[t]{0.27\textwidth}
            \centering
            \includegraphics[width=\linewidth]{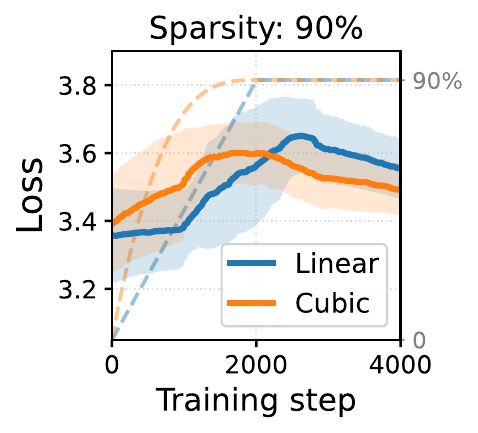}
        \end{subfigure}
        \setcounter{figure}{\value{figurebeforeappendixb}}
        \addtocounter{figure}{1}
        \captionof{figure}{Training-loss curves on OPT-125M under linear and cubic sparsity schedules (solid: train loss; dashed: sparsity, right axis).
        Cubic reaches lower final loss, with the gap widening at higher sparsity.}
        \label{fig:cubic_vs_linear_loss}
        \vspace{8pt}
        \end{@twocolumnfalse}%
    }
    \twocolumn[\warmupfiguretop]
    \setcounter{figure}{\value{figurebeforeappendixb}}
    \addtocounter{figure}{2}

    \begin{table}[!t]
    \centering
    \footnotesize
    \caption{Final perplexity on Wikitext-2 and C4 for OPT-125M under cubic vs linear sparsity schedule.}
    \label{tab:cubic_vs_linear}
    \renewcommand{\arraystretch}{1.05}
    \fontsize{7.5pt}{9pt}\selectfont
    \begin{tabular*}{0.92\linewidth}{@{\extracolsep{\fill}}l cc cc@{}}
    \toprule
    \multirow{3}{*}{Schedule}
    & \multicolumn{4}{c}{Sparsity} \\
    \cmidrule(lr){2-5}
    & \multicolumn{2}{c}{70\%}
    & \multicolumn{2}{c}{90\%} \\
    \cmidrule(lr){2-3}\cmidrule(lr){4-5}
    & Wiki & C4 & Wiki & C4 \\
    \midrule
    Cubic  & \textbf{30.77} & \textbf{29.16} & \textbf{43.84} & \textbf{39.15} \\
    Linear & 31.64 & 30.40 & 47.51 & 42.70 \\
    \bottomrule
    \end{tabular*}
    \end{table}

    \section{Ablation study}
    \label{appendix:ablation}
    \subsection{Warm-up steps}
    \label{appendix:warmup-choice}

    Here, we describe how the number of warm-up steps was chosen.
    Using OPT-125M with $T_{\text{total}}=4000$ training steps, we sweep the warm-up ratio $r \in \{0.1, 0.2, 0.3, 0.4, 0.5\}$ ($T_{\text{warmup}} = r \cdot T_{\text{total}} \in \{400, 800, 1200, 1600, 2000\}$) and report Wikitext-2 and C4 perplexity in \cref{fig:warmup_choice}.
    Across all sparsity levels, any warm-up is beneficial; longer warm-up, however, degrades performance because the sparsity schedule has already reached high sparsity while the learning rate is still ramping up, leaving insufficient capacity to recover from the aggressive pruning.
    This trend becomes more pronounced at higher sparsity.
    We therefore use a short warm-up of $r=0.1$ ($T_{\text{warmup}}=400$ steps) throughout the paper.

    \subsection{Cubic vs linear sparsity schedules}
    \label{appendix:cubic-vs-linear}
    Cubic sparsity schedules have been reported to be effective in prior pruning works.

    To check whether this still holds in our setting, we compare the cubic schedule adopted in \cref{subsec:schedule_design} (\cref{eq:cubic_schedule}) against the linear alternative defined as
    \begin{equation}
    S_t = S_{\text{init}} + (S_{\text{final}} - S_{\text{init}})\,\frac{t - t_0}{T_{\text{prune}}}.
    \label{eq:linear_schedule}
    \end{equation}

    On OPT-125M, with training-loss curves in \cref{fig:cubic_vs_linear_loss} and final perplexity in \cref{tab:cubic_vs_linear}, the cubic schedule introduces sparsity more aggressively in the early phase, causing a larger initial loss increase, but ends up at a lower final loss than the linear schedule, with the advantage most visible at 90\% sparsity.
    The same trend holds in perplexity: cubic wins at every sparsity level and the gap widens at higher sparsity.
    We therefore use the cubic schedule throughout the paper.

    \subsection{Additional results on the sparsity schedule}
    \label{appendix:no-schedule}

    \begin{figure*}[t]
        \centering
        \begin{subfigure}[b]{0.799\linewidth}
            \centering
            \includegraphics[width=\linewidth]{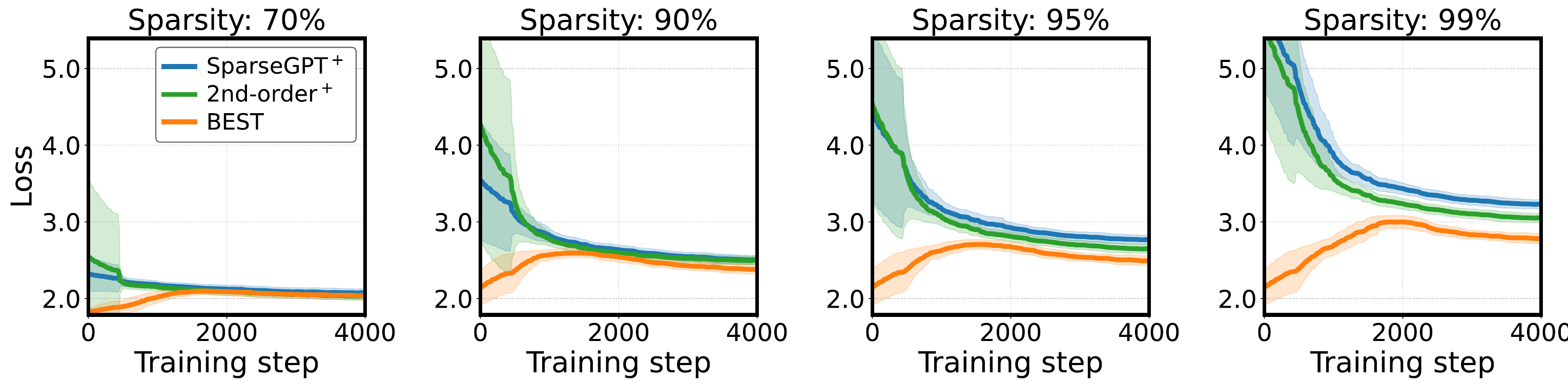}
            \caption{One-shot vs Progressive}
            \label{fig:no_schedule_curves_loss}
        \end{subfigure}%
        \hfill
        \begin{subfigure}[b]{0.1995\linewidth}
            \centering
            \includegraphics[width=\linewidth]{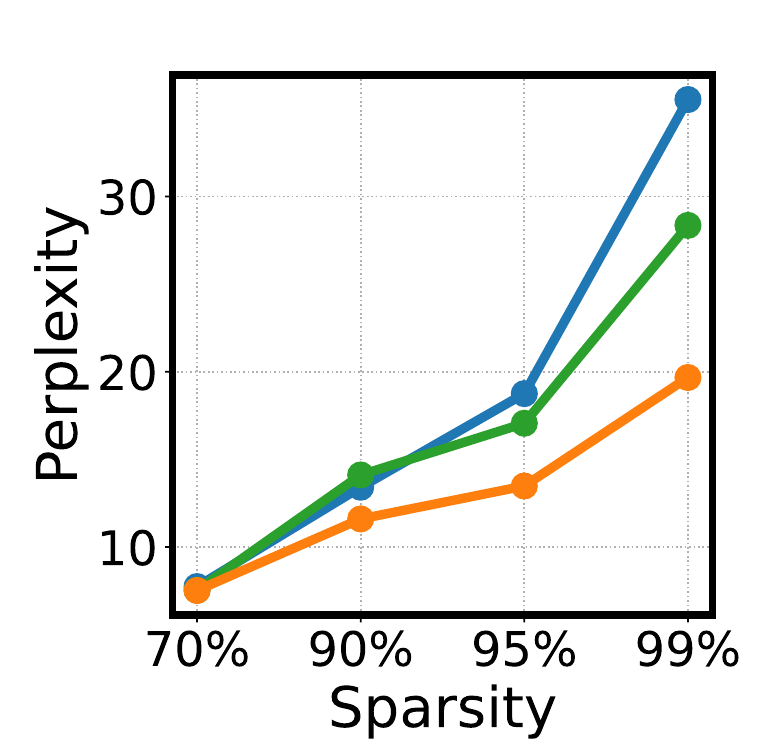}
            \caption{Perplexity}
            \label{fig:no_schedule_curves_ppl}
        \end{subfigure}
        \caption{Training loss curves on LLaMA-2-7B at each sparsity level, with the resulting WikiText-2 perplexity on the right.
        The comparison between \ours{} and 2nd-order\plus{} isolates the gradual schedule.
        The advantage of the gradual schedule grows with sparsity, with \ours{} pulling further ahead at each higher level.}
        \label{fig:no_schedule_curves}
    \end{figure*}

    \begin{table}[!t]
    \centering
    \footnotesize
    \begin{minipage}[t]{\linewidth}
        \centering
        \captionof{table}{Effect of reusing Adam optimizer state for diagonal Fisher approximation.
        $|D|$ denotes the calibration dataset size.}
        \label{tab:fisher_recompute_ablation}
        \renewcommand{\arraystretch}{1.05}
        \fontsize{7.5pt}{9pt}\selectfont
        \begin{tabular*}{0.92\linewidth}{@{\extracolsep{\fill}}c c cc cc@{}}
        \toprule
        \multirow{2}{*}{\emph{Recompute}} & \multirow{2}{*}{$|D|$}
        & \multicolumn{4}{c}{Sparsity} \\
        \cmidrule(lr){3-6}
        &
        & \multicolumn{2}{c}{70\%}
        & \multicolumn{2}{c}{90\%} \\
        &
        & Wiki & C4
        & Wiki & C4 \\
        \midrule
        \ding{55} & --   & \textbf{30.77} & 29.16 & \underline{43.84} & \textbf{39.15} \\
        \ding{51} & 128  & 31.32 & \underline{29.61} & 44.51 & 39.76 \\
         & 512 & \underline{30.96} & \textbf{29.54} & \textbf{43.79} & 39.39 \\
         & 2048 & 31.17 & 29.62 & 43.96 & \underline{39.24} \\
        \bottomrule
        \end{tabular*}
    \end{minipage}

    \par\vspace{0.8em}

    \begin{minipage}[t]{\linewidth}
        \centering
        \captionof{table}{Controlled comparison of global thresholding (\ours{}) and ATP layer-wise allocation on OPT-125M, with all other components fixed.}
        \label{tab:allocation_comparison}
        \renewcommand{\arraystretch}{1.05}
        \fontsize{7.5pt}{9pt}\selectfont
        \begin{tabular*}{0.92\linewidth}{@{\extracolsep{\fill}}l cc cc@{}}
            \toprule
            & \multicolumn{4}{c}{Sparsity} \\
            \cmidrule(lr){2-5}
            & \multicolumn{2}{c}{70\%} & \multicolumn{2}{c}{90\%} \\
            Allocation & Wiki & C4 & Wiki & C4 \\
            \midrule
            ATP & $31.75$ & $30.54$ & $45.28$ & $40.78$ \\
            \ours{} & $\mathbf{30.77}$ & $\mathbf{29.16}$ & $\mathbf{43.84}$ & $\mathbf{39.15}$ \\
            \bottomrule
        \end{tabular*}
    \end{minipage}
    \end{table}

    \Cref{subsec:ablation} demonstrates the benefit of progressive sparsification over one-shot pruning at $90\%$ sparsity.
    To further substantiate this finding, \cref{fig:no_schedule_curves} extends the comparison to all evaluated sparsity levels on LLaMA-2-7B and reports the resulting WikiText-2 perplexities.
    For reference, we also include SparseGPT\plus{} as a competitive retrained one-shot variant.
    At every evaluated sparsity level, \ours{} yields lower perplexity than both 2nd-order\plus{} and SparseGPT\plus{}, with the margins widening as sparsity increases.
    Specifically, from $70\%$ to $99\%$ sparsity, the gap grows from $0.01$ to $8.68$ against 2nd-order\plus{} and from $0.22$ to $15.86$ against SparseGPT\plus{}.
    These results indicate that progressive sparsification mitigates the limitations of imposing sparsity in a single step and becomes increasingly important at higher sparsity.

    \subsection{Diagonal Fisher approximation via Adam second moment}
    \label{appendix:statistics}
    
    We conduct ablation studies on OPT-125M to validate the effectiveness of reusing the Adam optimizer state for second-order saliency.
    To compute the saliency score for each parameter $w_i$, we contrast two strategies: directly reusing the accumulated second-moment element $v_{t,i}$ versus recomputing the empirical Fisher $\mathbb{E}[g_i^2]$ using gradients collected from 128, 512, and 2048 data samples.
    
    \cref{tab:fisher_recompute_ablation} presents the perplexity on Wikitext-2 and C4 datasets across sparsity levels of 70\% and 90\%.
    The results demonstrate that recomputing the second-moment estimate yields no consistent performance improvement over simple state reuse.
    Even with 2048 calibration sequences, the perplexity differences remain marginal across all sparsity regimes.
    Considering that recomputing $\mathbb{E}[g_i^2]$ introduces non-trivial computational overhead due to additional forward-backward passes, we conclude that reusing the existing Adam state provides a sufficiently accurate and efficient proxy for curvature in gradual pruning.
    
    \subsection{Alternative layer-wise sparsity allocation}
    \label{appendix:allocation}

    \cref{subsec:ablation} compares \ours{}'s global thresholding with uniform layer-wise allocation.
    To further validate the effectiveness of our allocation strategy, we replace only the allocation rule with ATP \citep{huang2025determining} on OPT-125M, retaining the saliency criterion, gradual schedule, optimization, and training budget of \ours{}.
    ATP assigns per-block sparsity targets along an arithmetic progression; at each target sparsity, we tune its common difference over a five-point grid spanning the feasible range.

    \cref{tab:allocation_comparison} shows that our global thresholding achieves lower perplexity than ATP at both sparsity levels on both corpora.
    These results demonstrate that our approach provides an effective sparsity allocation across layers in this setting.

    \def\additionaltablestopblock{%
    \begin{minipage}{\textwidth}
        \centering
        \footnotesize
        \captionof{table}{Wikitext-2 perplexity of models with comparable effective parameter counts.}
        \label{tab:table2}
        \renewcommand{\arraystretch}{1.0}
        \resizebox{0.35\linewidth}{!}{
        \begin{tabular}{l c c}
            \toprule
            Model & Sparsity & Perplexity \\
            \midrule
            OPT-125M & $0\%$ & $27.60$ \\
            OPT-1.3B & $90.5\%$ & $\mathbf{25.23}$ \\
            \midrule
            Qwen2.5-0.5B & $0\%$ & $12.87$ \\
            Qwen2.5-1.5B & $67\%$ & $\mathbf{12.03}$ \\
            \bottomrule
        \end{tabular}}

        \vspace{1em}

        \captionof{table}{Open-ended generation and reasoning accuracy (\%) for Qwen3-8B-Base pruned with \ours{}.
        Reasoning degrades far faster than perplexity or multiple-choice accuracy.}
        \label{tab:generation}
        \renewcommand{\arraystretch}{1.05}
        \resizebox{0.58\linewidth}{!}{
        \begin{tabular}{l c cccc}
            \toprule
            & \multirow{2}{*}{Dense} & \multicolumn{4}{c}{Sparsity} \\
            \cmidrule(lr){3-6}
            Benchmark & & $70\%$ & $90\%$ & $95\%$ & $99\%$ \\
            \midrule
            GSM8K (8-shot, CoT) & $87.64$ & $28.05$ & $2.35$ & $2.88$ & $2.12$ \\
            NQ-Open (5-shot)    & $25.9$  & $14.4$  & $5.6$  & $3.6$  & $1.0$ \\
            \bottomrule
        \end{tabular}}

        \vspace{1em}

        \captionof{table}{Representative GSM8K generations from Qwen3-8B-Base pruned with \ours{}.
        The $70\%$ model stays fluent but skips the overtime hours; the $95\%$ model collapses into repetition.}
        \label{tab:generation_example}
        \renewcommand{\arraystretch}{1.05}
        \begin{tabular}{@{}l p{0.88\linewidth}@{}}
            \toprule
            \multicolumn{2}{@{}p{0.95\linewidth}@{}}{\emph{Eliza earns \$10/hr for the first 40 hours and 1.2$\times$ that rate for overtime.
            She worked 45 hours.
            What are her earnings?} (answer: \$460)} \\
            \midrule
            Dense & ``$40 \times 10 = 400$.
            Overtime $1.2 \times 10 = \$12$/hr, $5 \times 12 = 60$.
            Total $400 + 60 = 460$.'' \\
            $70\%$ & ``$40 \times 10 = 400$.
            Overtime $1.2 \times 10 = \$12$.
            Total $400 + 12 = 412$.'' \\
            $95\%$ & ``Eliza's rate per hour for the first 40 hours\ldots{} If Eliza worked for 45 hours\ldots{} Eliza's rate\ldots'' \\
            \bottomrule
        \end{tabular}
    \end{minipage}
    \vspace{1em}
    }

    \def\figureeighttopblock{%
    \begin{minipage}{\textwidth}
        \centering
        \captionsetup{type=figure}
        \subcaptionbox{Sparsity allocation heatmaps for OPT-125M pruned to $90\%$ sparsity.\label{fig:sparsity_heatmap_90}}[\textwidth]{%
            \includegraphics[width=0.7\textwidth]{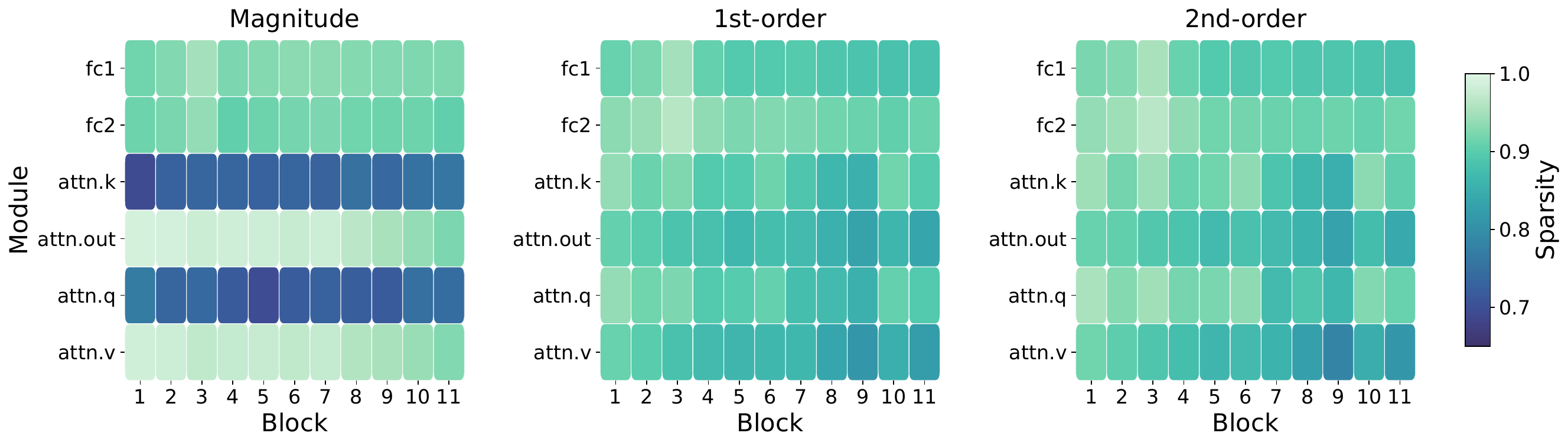}}\\[1em]
        \subcaptionbox{Per-block weight distribution of OPT-125M (KDE), drawn as a ridgeline with one row per Linear module type.
        Each panel corresponds to one Transformer block, and all panels share a single density scale so they are directly comparable.
        Module-level heterogeneity is most pronounced in early blocks and gradually flattens in later blocks.\label{fig:weight_dist_perblock}}[\textwidth]{%
            \includegraphics[width=0.7\textwidth]{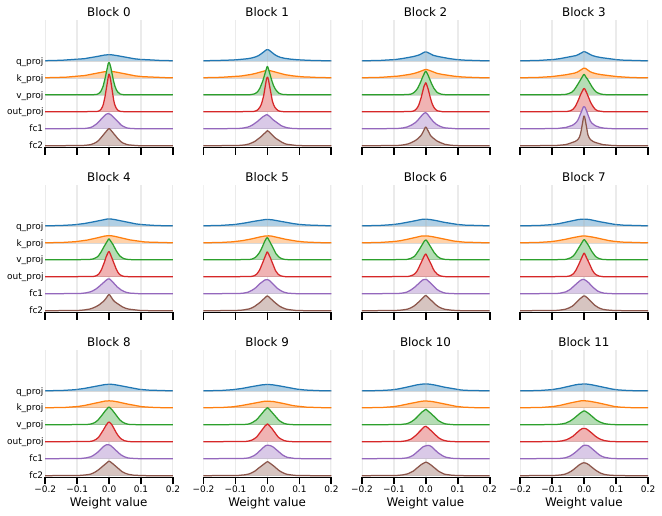}}
        \addtocounter{figure}{-1}%
        \captionof{figure}{Why magnitude pruning fails under global pruning.
        (a) Module-level sparsity allocation under magnitude vs.\ first-/second-order saliencies on OPT-125M at $90\%$ sparsity.
        (b) Block-wise weight distributions explain the allocation pattern.}
        \label{fig:why_magnitude_fails}
    \end{minipage}
    \vspace{1em}
    }
    \twocolumn[\figureeighttopblock]

    \begin{table}[!t]
        \centering
        \footnotesize
        \caption{WikiText-2 perplexity of LLaMA-2-7B across sparsity levels.
        \ours{}$^{\ddagger}$ denotes \ours{} under ELSA's training setting.}
        \label{tab:elsa_vs_best}
        \renewcommand{\arraystretch}{1.05}
        \fontsize{7.5pt}{9pt}\selectfont
        \begin{tabular*}{0.92\linewidth}{@{\extracolsep{\fill}}l cccc@{}}
            \toprule
            \multirow{2}{*}{Method} & \multicolumn{4}{c}{Sparsity} \\
            \cmidrule(lr){2-5}
            & $70\%$ & $90\%$ & $95\%$ & $99\%$ \\
            \midrule
            ELSA                 & $13.20$ & $26.97$ & $38.91$ & $55.94$ \\
            \ours{}$^{\ddagger}$ & $\underline{12.58}$ & $\underline{18.84}$ & $\underline{24.34}$ & $\underline{41.05}$ \\
            \ours{}             & $\mathbf{7.52}$ & $\mathbf{11.60}$ & $\mathbf{13.48}$ & $\mathbf{19.67}$ \\
            \bottomrule
        \end{tabular*}
    \end{table}

    \section{Additional results}
    \label{appendix:additional-results}

    \subsection{Extended comparison with ELSA}
    \label{appendix:elsa-comparison}

    We report the complete comparison with ELSA from $70\%$ to $99\%$ sparsity, including a matched variant of our method.
    Specifically, \ours{}$^{\ddagger}$ follows ELSA's exact training setting, using C4 and a $67$M-token budget.
    As shown in \cref{tab:elsa_vs_best}, \ours{}$^{\ddagger}$ outperforms ELSA at every sparsity level.
    Notably, at $95\%$ sparsity, \ours{}$^{\ddagger}$ even achieves lower perplexity than ELSA at $90\%$ sparsity ($24.34$ vs.\ $26.97$).

    Scaling the training budget to $524$M tokens and using SlimPajama, a cleaned and deduplicated pretraining corpus \citep{cerebras2023slimpajama}, further improves performance in the extreme-sparsity regime.
    Under this setting, \ours{} achieves perplexities of $13.48$ at $95\%$ sparsity and $19.67$ at $99\%$ sparsity.

    \begin{table*}[!t]
    \additionaltablestopblock
    \end{table*}

    \subsection{Why magnitude pruning fails: sparsity allocation and module-level weight distributions}
    \label{appendix:sparsity-allocation}
    \label{sec:why_magnitude_fails}

    In addition to the $70\%$ heatmap shown in \cref{subsec:ablation}, we report the per-module sparsity allocation of OPT-125M at $90\%$ (\cref{fig:sparsity_heatmap_90}).
    The same trend persists: magnitude pruning over-allocates sparsity to specific modules ($v_{\text{proj}}$ and $\text{out}_{\text{proj}}$) while leaving others ($q_{\text{proj}}$ and $k_{\text{proj}}$) relatively dense.
    This excessive allocation to a few modules is what causes magnitude pruning to fail under global pruning, as observed in \cref{tab:comparison_group} of \cref{subsec:ablation}.

    To explain the underlying cause, we analyze the weight distribution of every Transformer block in OPT-125M.
    \cref{fig:weight_dist_perblock} shows the empirical density of weight values per block, with each panel showing one row per Linear module type ($q_{\text{proj}}$, $k_{\text{proj}}$, $v_{\text{proj}}$, $\text{out}_{\text{proj}}$, $\text{fc}_1$, $\text{fc}_2$).
    Each block exhibits clear module-level heterogeneity, which is most severe in the early blocks and gradually flattens toward later blocks.
    As a result, as shown in \cref{fig:sparsity_heatmap_90}, magnitude pruning assigns disproportionately high sparsity to the early blocks of $v_{\text{proj}}$ and $\text{out}_{\text{proj}}$, leading to layer collapse.
    This indicates that the magnitude saliency score does not faithfully capture the actual loss contribution of each parameter.

    \subsection{Small dense vs.\ large sparse}
    \label{appendix:small-dense-vs-large-sparse}

    A practical question is whether \ours{} can convert an existing large pretrained model into compact models of different target sizes, rather than training a new small dense model from scratch for each size, which typically requires orders of magnitude more compute.
    We therefore compare small dense models with larger models pruned by \ours{} at comparable effective parameter counts, using this setting as a quality-oriented proxy for compact model construction.
    As shown in \cref{tab:table2}, the large sparse models achieve lower Wikitext-2 perplexity than their dense counterparts in both pairs.
    This trend holds across the OPT and Qwen2.5 families, suggesting that, when a strong pretrained model is already available, sparsification can provide a flexible and compute-efficient route to competitive compact models.
    By varying the target sparsity, \ours{} can produce models at different effective sizes without requiring a separate pretraining run for each target scale.
    We note that this comparison concerns model quality at a comparable effective parameter count; small dense models may still be preferable for deployment on hardware without efficient sparse computation support.

    \subsection{Open-ended generation and reasoning}
    \label{appendix:generation}

    To strengthen our assessment of the practical capabilities of sparse models, we extend the evaluation beyond perplexity and multiple-choice accuracy to open-ended generation and reasoning.
    Specifically, we evaluate Qwen3-8B-Base pruned with \ours{} on GSM8K (chain-of-thought, 8-shot) and NQ-Open (5-shot) using \texttt{lm-eval-harness}.
    These evaluations use the pruned base models without task-specific data curation or adaptation.

    \cref{tab:generation} shows that open-ended generation and reasoning remain challenging as sparsity increases.
    Mathematical reasoning degrades fastest: GSM8K falls from $87.64$ to $28.05$ already at $70\%$ sparsity and to near-zero accuracy beyond it, while factual recall on NQ-Open decays more gradually.
    \cref{tab:generation_example} illustrates this failure mode: at $70\%$, the model still produces well-formed arithmetic but omits a step of the problem, whereas at $95\%$, it stops solving and repeats fragments of the question.
    These findings show that preserving perplexity and multiple-choice accuracy under extreme sparsity does not ensure robust free-form generation or reasoning.
    Recovering these capabilities through sparsity-aware adaptation or targeted recovery training is therefore an important direction for future work.

    \subsection{Numerical results of zero-shot downstream task accuracy}
    \label{appendix:zero-shot}
    \label{appendix:zs-llama-7b-detail}

    We further provide the full per-task zero-shot results for LLaMA-2-7B/13B and Qwen3-4B/8B-Base, as discussed in \cref{sec:exp}.

    \clearpage

    \begin{table*}[!ht]
        \centering
        \fontsize{7.5pt}{9pt}\selectfont
        \caption{Zero-shot accuracy (\%) of LLaMA-2-7B across multiple tasks under different sparsity regimes.
        Numbers for L-ADMM, ALPS, SAFE, SparseLLM, and ELSA are taken from \citet{lee2025unseen}.}
        \label{tab:zeroshot_llama2}
        \begin{tabular}{c l cccccccc}
            \toprule
                \textbf{Sparsity} & \textbf{Method}
                & \textbf{ARC-C} & \textbf{ARC-E} & \textbf{BoolQ}
                & \textbf{HellaSwag} & \textbf{OBQA} & \textbf{RTE}
                & \textbf{Winogrande} & \textbf{Avg} \\
            \midrule

            0\% & Dense
            & 43.35 & 76.26 & 77.68 & 57.14 & 31.40 & 62.82 & 69.06 & 59.67 \\
            \midrule

            \multirow{12}{*}{70\%}
                & Magnitude      & 22.70 & 27.82 & 37.95 & 25.92 & 17.00 & 53.07 & 49.09 & 33.36 \\
                & Wanda          & 18.52 & 30.93 & 47.65 & 28.11 & 13.20 & 52.71 & 49.01 & 34.30 \\
                & SparseGPT      & 21.84 & 44.61 & 65.02 & 34.53 & 16.80 & 52.71 & 59.12 & 42.09 \\
                & L-ADMM         & 23.81 & 50.63 & 63.21 & 36.57 & 20.40 & 54.15 & 60.77 & 44.22 \\
                & ALPS           & 25.51 & 52.78 & 63.46 & 37.54 & 20.8  & 53.43 & 61.72 & 45.03 \\
                & SAFE           & 24.23 & 45.62 & 43.76 & 34.74 & 18.40 & 52.71 & 53.12 & 38.94 \\
                & SparseLLM      & 20.90 & 40.32 & 61.87 & 32.74 & 16.0  & 54.51 & 57.46 & 40.54 \\
                & ELSA           & 27.13 & 55.81 & 63.61 & 43.16 & 22.40 & 52.71 & 58.64 & 46.21 \\
                & Magnitude\plus{}  & 32.08 & \underline{66.41} & \underline{68.01} & 48.73 & \textbf{26.80} & \textbf{59.93} & \underline{63.30} & \textbf{52.18} \\
                & Wanda\plus{}      & 30.55 & 63.68 & 65.11 & 46.45 & 22.60 & \underline{57.04} & 60.06 & 49.36 \\
                & SparseGPT\plus{}  & \underline{32.94} & 64.23 & \textbf{69.24} & \underline{48.47} & 25.40 & 53.07 & \textbf{63.69} & 51.00 \\
                & \ours           & \textbf{34.30} & \textbf{68.77} & 67.98 & \textbf{49.67} & \underline{25.60}  & 55.57 & 62.51 & \underline{52.06} \\
            \midrule
            \multirow{12}{*}{90\%}
                & Magnitude
                    & \textbf{22.53} & 25.88 & 39.17 & 25.50 & 16.00 & 47.29 & 49.96 & 32.33 \\
                & Wanda
                    & 20.90 & 26.05 & 37.83 & 25.76 & 15.20 & 52.71 & 49.88 & 32.62 \\
                & SparseGPT
                    & 21.25 & 25.84 & 38.04 & 25.59 & 12.60 & 53.43 & 49.88 & 32.38 \\
                & L-ADMM
                    & 19.97 & 26.14 & 37.83 & 26.46 & 13.60 & 51.62 & 47.51 & 31.88 \\
                & ALPS
                    & 19.45 & 26.89 & 37.8  & 26.81 & 12.8 & \textbf{53.79} & 46.65 & 32.03 \\
                & SAFE
                    & \underline{21.84} & 26.52 & 37.83 & 25.91 & 15.80 & 52.71 & 47.83 & 32.63 \\
                & SparseLLM
                    & 20.56 & 25.72 & 37.83 & 25.94 & 13.8 & 52.71 & 46.96 & 31.93 \\
                & ELSA
                    & 18.52 & 41.33 & 57.25 & 31.54 & 16.60 & 52.71 & 51.70 & 38.52 \\
                & Magnitude\plus{}
                    & 20.48 & \underline{48.27} & 58.07 & 33.82 & \underline{19.20} & 52.71 & 51.54 & \underline{40.58}  \\
                & Wanda\plus{}
                    & 18.94 & 42.63 & 58.78 & 30.49 & 15.40 & \underline{53.07} & 49.09 & 38.34  \\
                & SparseGPT\plus{}
                    & 19.03 & 45.12 & \underline{59.69} & \underline{31.86} & 16.40 & 52.71 & \underline{52.17} & 39.57 \\
                & \ours
                    & 20.48 & \textbf{49.49} & \textbf{60.95} & \textbf{36.51} & \textbf{22.4} & 52.71 & \textbf{53.83}  & \textbf{42.34}  \\
                \midrule

            \multirow{7}{*}{95\%}
                & Magnitude          & \textbf{22.01} & 25.97 & 52.05 & 25.64 & 14.60 & 45.13 & 49.80 & 33.60 \\
                & Wanda              & \underline{21.59} & 25.72 & 37.83 & 25.78 & 13.60 & 52.71 & 49.57 & 32.40 \\
                & SparseGPT          & 21.42 & 26.89 & 37.83 & 25.64 & 15.80 & \textbf{53.07} & 51.30 & 33.14 \\
                & Magnitude\plus{}   & 18.09 & 39.31 & 60.64 & 28.15 & 15.20 & 52.35 & \underline{51.38} & 37.87 \\
                & Wanda\plus{}       & 17.58 & 36.9  & \underline{61.65} & 27.94 & 15.00 & 52.71 & \textbf{52.25} & 37.72  \\
                & SparseGPT\plus{}   & 17.92 & \underline{39.35} & \textbf{61.83} & \underline{28.45} & \underline{16.60} & 52.35 & 49.09 & \underline{37.94} \\
                & \ours & 19.63 & \textbf{46.63} & 57.83 & \textbf{32.72} & \textbf{17.80} & 52.71 & 50.67 & \textbf{39.71}  \\
            \midrule

            \multirow{7}{*}{99\%}
                & Magnitude          & 21.67 & 26.52 & 51.71 & 25.36 & \underline{15.20} & 48.74 & 49.57 & 34.11 \\
                & Wanda              & \textbf{23.21} & 25.17 & \textbf{62.17} & 25.62 & \textbf{16.60} & 45.85 & 47.59 & 35.17 \\
                & SparseGPT          & \underline{22.18} & 25.34 & 41.90 & 25.65 & 14.40 & 46.21 & 50.12 & 32.26 \\
                & Magnitude\plus{}   & 16.98 & 34.39 & \underline{61.96} & 26.95 & 11.60 & 51.26 & \textbf{52.25} & 36.48 \\
                & Wanda\plus{}       & 18.34 & \underline{34.47} & 61.53 & 26.91 & 14.60  & \underline{53.07} & 51.54 & \underline{37.21}  \\
                & SparseGPT\plus{}   & 17.58 & 33.67 & 60.09 & \underline{27.27} & 13.20 & \textbf{53.79} & 50.51 & 36.59
                  \\
                & \ours & 17.58 & \textbf{40.19} & 61.41 & \textbf{28.43} & 13.60 & 53.07 & 50.59 & \textbf{37.84} \\
            \bottomrule
        \end{tabular}
    \end{table*}

    \clearpage

    \begin{table*}[!ht]
        \centering
        \fontsize{7.5pt}{9pt}\selectfont
        \caption{Zero-shot accuracy (\%) of LLaMA-2-13B across multiple tasks under different sparsity regimes.}
        \label{tab:zeroshot_llama2_13b}
        \begin{tabular}{c l cccccccc}
            \toprule
                \textbf{Sparsity} & \textbf{Method}
                & \textbf{ARC-C} & \textbf{ARC-E} & \textbf{BoolQ}
                & \textbf{HellaSwag} & \textbf{OBQA} & \textbf{RTE}
                & \textbf{Winogrande} & \textbf{Avg} \\
            \midrule
    
            0\% & Dense
            & 48.46 & 79.38 & 80.55 & 60.04 & 35.20 & 65.34 & 72.14 & 63.02 \\
            \midrule
    
            \multirow{7}{*}{70\%}
                & Magnitude        & 20.65 & 31.27 & 38.65 & 27.53 & 14.60 & 52.71 & 49.33 & 33.54 \\
                & Wanda            & 18.94 & 40.15 & 62.17 & 29.96 & 14.60 & 52.71 & 52.72 & 38.75 \\
                & SparseGPT        & 24.74 & 52.82 & 68.10 & 36.73 & 19.80 & 52.71 & 61.25 & 45.16 \\
                & Magnitude\plus{} & 36.26 & \underline{70.45} & 71.35 & 52.85 & 28.20 & \underline{61.37} & \underline{66.14} & \underline{55.23} \\
                & Wanda\plus{}     & 34.39 & 68.43 & 71.04 & 51.75 & 27.20 & 57.76 & 63.06 & 53.38 \\
                & SparseGPT\plus{} & \underline{36.60} & 69.99 & \textbf{72.78} & \underline{53.13} & \underline{28.60} & 59.21 & \textbf{66.30} & \underline{55.23} \\
                & \ours            & \textbf{37.63} & \textbf{72.69} & \underline{71.59} & \textbf{54.50} & \textbf{31.60} & \textbf{62.45} & 65.19 & \textbf{56.52} \\
            \midrule

            \multirow{7}{*}{90\%}
                & Magnitude        & 21.67 & 24.83 & 44.19 & 25.71 & 15.00 & 45.85 & 52.01 & 32.75 \\
                & Wanda            & 22.10 & 25.55 & 37.83 & 25.80 & 13.40 & 52.71 & 52.80 & 32.88 \\
                & SparseGPT        & 22.18 & 25.88 & 56.82 & 25.92 & 13.00 & 52.71 & 51.07 & 35.37 \\
                & Magnitude\plus{} & \underline{22.78} & \underline{52.19} & 58.29 & \underline{36.98} & \underline{21.60} & 52.71 & \underline{53.43} & \underline{42.57} \\
                & Wanda\plus{}     & 17.75 & 43.14 & \underline{60.06} & 31.48 & 14.60 & 52.71 & 51.62 & 38.76 \\
                & SparseGPT\plus{} & 21.33 & 48.78 & 58.38 & 34.49 & 17.00 & \underline{53.07} & 51.54 & 40.65 \\
                & \ours            & \textbf{25.43} & \textbf{56.73} & \textbf{62.14} & \textbf{40.78} & \textbf{23.00} & 52.71 & \textbf{57.06} & \textbf{45.41} \\
            \midrule
    
            \multirow{7}{*}{95\%}
                & Magnitude        & \textbf{24.06} & 25.63 & 58.23 & 25.48 & 15.40 & 47.29 & 48.62 & 34.96 \\
                & Wanda            & \underline{22.27} & 25.80 & 37.83 & 25.79 & 13.00 & 52.71 & 51.14 & 32.65 \\
                & SparseGPT        & 21.76 & 25.72 & 37.83 & 25.76 & 14.60 & 52.71 & 50.12 & 32.64 \\
                & Magnitude\plus{} & 17.75 & 33.75 & \underline{61.96} & 26.77 & 12.80 & 52.71 & 50.67 & 36.63 \\
                & Wanda\plus{}     & 17.06 & 39.65 & 61.74 & 28.19 & \underline{16.00} & 52.71 & 50.91 & 38.04 \\
                & SparseGPT\plus{} & 17.83 & \underline{40.32} & 61.47 & \underline{29.35} & 13.80 & \textbf{53.07} & \underline{52.72} & \underline{38.37} \\
                & \ours            & 20.56 & \textbf{48.70} & \textbf{61.99} & \textbf{34.46} & \textbf{18.60} & 52.71 & \textbf{53.51} & \textbf{41.50} \\
            \midrule

            \multirow{7}{*}{99\%}
                & Magnitude        & \textbf{22.78} & 25.21 & 52.72 & 25.61 & \underline{14.80} & 51.99 & 49.96 & 34.72 \\
                & Wanda            & \underline{22.35} & 25.38 & 39.30 & 25.75 & 13.60 & 47.29 & 50.12 & 31.97 \\
                & SparseGPT        & 21.67 & 25.63 & 43.27 & 25.65 & 14.00 & \underline{53.79} & 48.93 & 33.28 \\
                & Magnitude\plus{} & 17.75 & 33.67 & 60.43 & 26.87 & 13.40 & 52.71 & 49.49 & 36.33 \\
                & Wanda\plus{}     & 17.92 & \underline{34.68} & \underline{60.95} & 27.03 & 13.00 & 52.71 & \underline{50.59} & 36.70 \\
                & SparseGPT\plus{} & 17.41 & 34.64 & \textbf{62.23} & \underline{27.33} & 12.60 & \textbf{54.15} & \textbf{52.17} & \textbf{37.22} \\
                & \ours            & 18.09 & \textbf{36.53} & 59.57 & \textbf{27.95} & 15.20 & 53.07 & 50.12 & \textbf{37.22} \\
            \bottomrule
    \end{tabular}
    \end{table*}

    \begin{table*}[!ht]
        \centering
        \fontsize{7.5pt}{9pt}\selectfont
        \caption{Zero-shot accuracy (\%) of Qwen3-4B-Base across multiple tasks under different sparsity regimes.}
        \label{tab:zeroshot_qwen3_4b}
        \begin{tabular}{c l cccccccc}
            \toprule
                \textbf{Sparsity} & \textbf{Method}
                & \textbf{ARC-C} & \textbf{ARC-E} & \textbf{BoolQ}
                & \textbf{HellaSwag} & \textbf{OBQA} & \textbf{RTE}
                & \textbf{Winogrande} & \textbf{Avg} \\
            \midrule
    
            0\% & Dense
            & 48.38 & 79.17 & 83.12 & 54.50 & 31.60 & 76.90 & 70.56 & 63.46 \\
            \midrule
    
            \multirow{7}{*}{70\%}
                & Magnitude        & 22.18 & 25.29 & 49.27 & 25.69 & 16.80 & 53.43 & 50.59 & 34.75 \\
                & Wanda            & 19.03 & 43.22 & 62.14 & 29.31 & 12.00 & 52.71 & 50.67 & 38.44 \\
                & SparseGPT        & 25.68 & 55.18 & 62.26 & 35.71 & 17.80 & 54.51 & 57.14 & 44.04 \\
                & Magnitude\plus{} & 35.58 & 69.57 & 70.40 & 45.48 & 26.40 & \textbf{59.93} & 61.17 & 52.65 \\
                & Wanda\plus{}     & 34.13 & 69.57 & 67.19 & 44.85 & 26.20 & 57.76 & 58.96 & 51.24 \\
                & SparseGPT\plus{} & \underline{37.80} & \underline{69.99} & \underline{70.80} & \underline{46.17} & \underline{26.80} & \underline{58.84} & \textbf{62.12} & \underline{53.22} \\
                & \ours            & \textbf{39.16} & \textbf{73.44} & \textbf{73.21} & \textbf{47.73} & \textbf{27.80} & 57.76 & \underline{61.25} & \textbf{54.34} \\
            \midrule
    
            \multirow{7}{*}{90\%}
                & Magnitude        & 19.88 & 24.49 & \underline{60.18} & 25.95 & 14.80 & 46.93 & \underline{52.57} & 34.97 \\
                & Wanda            & 20.56 & 24.62 & 37.86 & 25.59 & 14.20 & 51.99 & 49.09 & 31.99 \\
                & SparseGPT        & 20.05 & 26.89 & 38.23 & 26.21 & 11.20 & \textbf{52.71} & 48.78 & 32.01 \\
                & Magnitude\plus{} & 17.75 & 43.27 & 60.06 & 28.99 & 16.40 & 50.90 & 50.43 & 38.26 \\
                & Wanda\plus{}     & 20.31 & 44.40 & 58.32 & 29.59 & 13.40 & 52.35 & 50.75 & 38.45 \\
                & SparseGPT\plus{} & \underline{20.56} & \underline{46.42} & \textbf{61.68} & \underline{31.27} & \underline{16.80} & 52.35 & 50.75 & \underline{39.83} \\
                & \ours            & \textbf{22.78} & \textbf{53.32} & 60.09 & \textbf{36.42} & \textbf{20.80} & \textbf{52.71} & \textbf{54.78} & \textbf{42.99} \\
            \midrule
    
            \multirow{7}{*}{95\%}
                & Magnitude        & \underline{20.73} & 24.41 & 38.53 & 25.57 & 15.60 & 48.38 & 49.41 & 31.80 \\
                & Wanda            & \textbf{21.42} & 26.47 & 37.83 & 25.76 & 14.00 & 52.71 & 48.07 & 32.32 \\
                & SparseGPT        & 20.56 & 26.09 & 38.07 & 25.95 & \textbf{19.40} & 52.71 & 49.88 & 33.24 \\
                & Magnitude\plus{} & 18.60 & 37.21 & 60.76 & 27.18 & 13.00 & \textbf{53.43} & \underline{51.38} & 37.37 \\
                & Wanda\plus{}     & 18.17 & \underline{39.90} & 59.02 & \underline{28.08} & 14.00 & \textbf{53.43} & 50.04 & 37.52 \\
                & SparseGPT\plus{} & 17.49 & 39.81 & \textbf{62.17} & 27.74 & 13.20 & 52.71 & 50.20 & \underline{37.62} \\
                & \ours            & 20.65 & \textbf{49.24} & \underline{61.10} & \textbf{32.73} & \underline{17.40} & \underline{53.07} & \textbf{51.62} & \textbf{40.83} \\
            \midrule
    
            \multirow{7}{*}{99\%}
                & Magnitude        & \underline{21.84} & 25.25 & 38.56 & 25.56 & \underline{16.00} & 53.07 & 50.51 & 32.97 \\
                & Wanda            & \textbf{24.83} & 24.62 & 38.23 & 25.15 & \textbf{16.20} & 52.71 & 50.75 & 33.20 \\
                & SparseGPT        & \underline{21.84} & 24.37 & 37.80 & 25.79 & 13.20 & 52.35 & \underline{51.38} & 32.39 \\
                & Magnitude\plus{} & 18.34 & 33.67 & 45.57 & 26.48 & 13.60 & 51.62 & 51.14 & 34.35 \\
                & Wanda\plus{}     & 17.15 & \underline{35.10} & 54.46 & \underline{26.70} & 11.80 & \textbf{54.87} & 50.43 & 35.79 \\
                & SparseGPT\plus{} & 18.52 & 34.34 & \underline{61.01} & 26.44 & 13.40 & \underline{53.43} & 50.12 & \underline{36.75} \\
                & \ours            & 18.52 & \textbf{40.11} & \textbf{61.80} & \textbf{28.50} & 12.80 & 52.35 & \textbf{52.09} & \textbf{38.02} \\
            \bottomrule
    \end{tabular}
    \end{table*}

    \begin{table*}[!ht]
        \centering
        \fontsize{7.5pt}{9pt}\selectfont
        \caption{Zero-shot accuracy (\%) of Qwen3-8B-Base across multiple tasks under different sparsity regimes.}
        \label{tab:zeroshot_qwen3}
        \begin{tabular}{c l cccccccc}
            \toprule
                \textbf{Sparsity} & \textbf{Method}
                & \textbf{ARC-C} & \textbf{ARC-E} & \textbf{BoolQ}
                & \textbf{HellaSwag} & \textbf{OBQA} & \textbf{RTE}
                & \textbf{Winogrande} & \textbf{Avg} \\
            \midrule
    
            0\% & Dense
            & 52.56 & 81.69 & 83.09 & 58.80 & 32.40 & 74.73 & 72.38 & 65.09 \\
            \midrule
    
            \multirow{7}{*}{70\%}
                & Magnitude        & 21.84 & 27.48 & 44.31 & 25.84 & 17.20 & 53.43 & 49.96 & 34.30 \\
                & Wanda            & 20.31 & 48.99 & 62.17 & 30.47 & 16.00 & 52.71 & 53.67 & 40.62 \\
                & SparseGPT        & 29.78 & 62.04 & 65.66 & 38.75 & 21.80 & 58.84 & 60.69 & 48.22 \\
                & Magnitude\plus{} & 42.06 & \underline{75.80} & 72.17 & 49.63 & \underline{28.20} & 63.90 & 62.83 & 56.37 \\
                & Wanda\plus{}     & \underline{42.49} & 74.49 & 71.65 & 49.05 & 27.20 & 62.09 & 63.30 & 55.75 \\
                & SparseGPT\plus{} & 42.24 & 74.79 & \underline{73.03} & \underline{50.35} & \textbf{30.40} & \textbf{67.15} & \underline{64.01} & \underline{57.42} \\
                & \ours            & \textbf{44.80} & \textbf{76.35} & \textbf{73.36} & \textbf{51.17} & 27.80 & \underline{64.62} & \textbf{66.69} & \textbf{57.83} \\
            \midrule
    
            \multirow{7}{*}{90\%}
                & Magnitude        & 22.10 & 25.72 & 43.03 & 26.00 & 15.00 & 51.26 & 48.78 & 33.13 \\
                & Wanda            & 20.14 & 24.79 & 37.77 & 26.15 & 12.60 & 52.35 & 49.25 & 31.86 \\
                & SparseGPT        & 20.22 & 27.15 & 44.40 & 26.57 & 12.60 & 51.99 & 50.12 & 33.29 \\
                & Magnitude\plus{} & \underline{22.35} & 47.69 & 60.43 & 31.71 & 18.60 & \underline{54.15} & 51.62 & 40.93 \\
                & Wanda\plus{}     & 20.39 & 47.35 & 60.89 & 32.18 & 18.40 & 52.35 & 50.59 & 40.31 \\
                & SparseGPT\plus{} & 21.50 & \underline{49.83} & \underline{61.28} & \underline{34.06} & \underline{19.00} & \textbf{55.23} & \underline{53.28} & \underline{42.03} \\
                & \ours            & \textbf{25.00} & \textbf{56.48} & \textbf{61.71} & \textbf{38.87} & \textbf{21.60} & 53.43 & \textbf{56.43} & \textbf{44.79} \\
            \midrule
    
            \multirow{7}{*}{95\%}
                & Magnitude        & 21.59 & 25.46 & 40.55 & 25.72 & \underline{16.80} & \textbf{52.71} & 48.30 & 33.02 \\
                & Wanda            & \textbf{23.46} & 25.93 & 54.56 & 25.72 & 16.60 & 52.35 & 49.33 & 35.42 \\
                & SparseGPT        & 20.22 & 28.03 & 37.83 & 25.77 & 13.20 & \textbf{52.71} & 49.64 & 32.48 \\
                & Magnitude\plus{} & 16.98 & 37.96 & \textbf{62.17} & 27.54 & 13.00 & 52.35 & 50.43 & 37.21 \\
                & Wanda\plus{}     & 17.75 & 40.95 & \underline{61.71} & 28.49 & 13.80 & \textbf{52.71} & \underline{51.14} & 38.08 \\
                & SparseGPT\plus{} & 17.75 & \underline{42.05} & \textbf{62.17} & 29.15 & 13.40 & 52.35 & 50.91 & \underline{38.25} \\
                & \ours            & \underline{23.04} & \textbf{50.76} & 59.76 & \textbf{34.91} & \textbf{18.60} & \textbf{52.71} & \textbf{51.30} & \textbf{41.58} \\
            \midrule
    
            \multirow{7}{*}{99\%}
                & Magnitude        & \underline{23.29} & 25.51 & 37.83 & 25.52 & \textbf{17.00} & \textbf{52.71} & 50.75 & 33.23 \\
                & Wanda            & \textbf{23.38} & 25.38 & 54.95 & 25.63 & 15.00 & \underline{51.99} & 49.96 & 35.18 \\
                & SparseGPT        & 22.35 & 25.46 & 38.10 & 25.45 & \underline{15.80} & 47.29 & 49.17 & 31.95 \\
                & Magnitude\plus{} & 16.47 & 36.62 & 59.48 & 27.15 & 12.00 & 50.90 & 49.96 & 36.08 \\
                & Wanda\plus{}     & 17.58 & 37.12 & 50.00 & 27.11 & 14.00 & 51.26 & 50.91 & 35.43 \\
                & SparseGPT\plus{} & 16.38 & \underline{38.51} & \textbf{62.02} & \underline{27.61} & 12.40 & 51.62 & 50.83 & \underline{37.05} \\
                & \ours            & 19.62 & \textbf{42.21} & \underline{61.87} & \textbf{29.49} & 15.60 & \textbf{52.71} & \textbf{51.46} & \textbf{38.99} \\
            \bottomrule
    \end{tabular}
    \end{table*}

    \clearpage

    \end{appendix}

\end{document}